\documentclass[journal]{IEEEtran}  

\usepackage{graphicx} 
\usepackage{amsmath}
\usepackage{amssymb}  
\usepackage{algorithm, algorithmic}
\newtheorem{theorem}{Theorem}
\newtheorem{lemma}[theorem]{Lemma}	

\newtheorem{remark}[theorem]{Remark}
\newtheorem{example}[theorem]{Example}	

\newtheorem{proposition}[theorem]{Proposition}	
	
\newtheorem{definition}[theorem]{Definition}	

\usepackage{color}
										
\usepackage[hidelinks]{hyperref} 

\newcommand{\clr}[1]{\textcolor{black}{#1}}

\newcommand{\rev}[1]{\textcolor{black}{#1}}

\newcommand\copyrighttext{%
	\footnotesize \textcopyright 2021 IEEE. Personal use of this material is permitted.  Permission from IEEE must be obtained for all other uses, in any current or future media, including reprinting/republishing this material for advertising or promotional purposes, creating new collective works, for resale or redistribution to servers or lists, or reuse of any copyrighted component of this work in other works.}
\usepackage{tikz}
\newcommand\copyrightnotice{%
	\begin{tikzpicture}[remember picture,overlay]
	\node[anchor=south,yshift=7pt] at (current page.south) {\fbox{\parbox{\dimexpr\textwidth-\fboxsep-\fboxrule\relax}{\copyrighttext}}};
	\end{tikzpicture}%
}

\title{Stabilization of Complementarity Systems via Contact-Aware Controllers
}

\author{Alp Aydinoglu$^{1}$, Philip Sieg$^{1}$, Victor M. Preciado$^{1}$, Michael Posa$^{1}$
\thanks{*This work was supported by the National Science Foundation under Grant No. CMMI-1830218.
}
    \thanks{$^{1}$Alp Aydinoglu, Philip Sieg and Michael Posa are with the General Robotics, Automation, Sensing and Perception (GRASP) Laboratory, Victor M. Preciado is with the Department of Electrical and Systems Engineering, University of Pennsylvania, Philadelphia, PA 19104, USA.
   {\tt\small \{alpayd, philsieg, preciado, posa\}@seas.upenn.edu} (\textit{Corresponding author: Alp Aydinoglu})}%

}

\begin{document}
\maketitle
\copyrightnotice


\begin{abstract}
We propose a control framework which can utilize tactile information by exploiting the complementarity structure of contact dynamics.	
Since many robotic tasks, like manipulation and locomotion, are fundamentally based in making and breaking contact with the environment, state-of-the-art control policies struggle to deal with the hybrid nature of multi-contact motion. 
Such controllers often rely heavily upon heuristics or, due to the \clr{combinatorial} structure in the dynamics, are unsuitable for real-time control.
Principled deployment of tactile sensors offers a promising mechanism for stable and robust control, but modern approaches often use this data in an ad hoc manner, for instance to guide guarded moves.
This framework can close the loop on tactile sensors and 
it is \clr{non-combinatorial}, enabling optimization algorithms to automatically synthesize provably stable control policies.
\clr{ We demonstrate this approach on multiple numerical examples, including quasi-static friction problems and a high dimensional problem with ten contacts. We also validate our results on an experimental setup and show the effectiveness of the proposed method on an underactuated multi-contact system.}
\end{abstract}

\begin{IEEEkeywords}
	Tactile feedback, optimization-based control, bilinear matrix inequalities, force control
\end{IEEEkeywords}

\IEEEpeerreviewmaketitle

\section{Introduction}
\IEEEPARstart{I}{n} recent years, robotic automation has excelled in dealing with repetitive tasks in static and structured environments.
\clr{ On the other hand, to achieve the promise of the field, robots must perform efficiently in complex, unstructured environments that involve physical interaction between the robot and the environment itself which has been an ongoing research direction for many years \cite{mills1993control,mills1993stability}}.
Furthermore, as compared with traditional motion planning problems, tasks like dexterous manipulation and legged locomotion fundamentally require intentionally initiating contact with the environment to achieve a positive result.
To enable stable, and robust motion, it is critically important to design policies that explicitly consider the interaction between robot and environment.

Contact, however, is hybrid or multi-modal in nature, capturing the effect of stick-slip transitions or making and breaking contact.
Standard approaches to control often match the hybrid dynamics with a hybrid or switching controller, where one policy is associated with each mode.
However, precise identification of the hybrid events is difficult in practice, and switching controllers can be brittle, particularly local to the switching surface, or require significant hand-tuning.
Model predictive control, closely related to this work, is one approach that has been regularly applied to control through contact, with notable successes.
\clr{
Due to the computational complexity of hybrid model predictive control, most of these approaches have not been demonstrated to work in real-time for dynamic problems \cite{mastalli2020crocoddyl}, \cite{winkler2018gait}. Methods
that work in real-time must either approximate the hybrid dynamics (e.g. \cite{tassa2012synthesis}), or limit online control to a known mode sequence \cite{Hogan2017}. Most of these approaches do not provide stability guarantees. Methods that provide guarantees \cite{marcucciwarm} require significant online computation time and have not been shown to work in real-time applications.
}

\clr{While prior work has explored computational synthesis of non-switching feedback policies \cite{posa2015stability}, it does not incorporate tactile sensing, and there are clear structural limits to smooth, non-switching, state-based control.}
Here, we focus on offline synthesis of a stabilizing feedback policy, eliminating the need for intensive online calculations.

The need for contact-aware control is driven, in part, by recent advances in tactile sensing (e.g \cite{wettels2009multi, guggenheim2017robust, yuan2015measurement, kumar2016optimal, donlon2018gelslim} and others).
Given these advances, there has been ongoing research to design control policies using tactile feedback for tasks that require making and breaking contact. 
However, these approaches are largely based on static assumptions, for instance with guarded moves \cite{howe1993tactile}, or rely upon switching controllers (e.g. \cite{romano2011human, yamaguchi2016combining}).
Other recent methods  incorporate tactile sensors within deep learning frameworks, though they offer no guarantees on performance or stability  \cite{merzic2018leveraging, tian2019manipulation}.

\clr{In this work, we present an optimization-based numerical approach for designing control policies that use feedback on the contact forces.}
The control policy combines regular state feedback with tactile feedback in order to provably stabilize systems with possibly non-unique solutions.
Our controller structure is \clr{non-combinatorial} in nature and avoids enumerating the exponential number of potential hybrid modes that might arise from contact.
\rev{More precisely, the contributions of each contact are additive, rather than combinatorial.}
Inspired by both prior work \cite{posa2015stability} and \cite{camlibel2001complementarity}, we synthesize and verify a corresponding non-smooth, piecewise quadratic Lyapunov function.
Additionally, we are able to explicitly define sparsity patterns allowing us to design controllers for systems where the full state information might be lacking, such as when the state of an object is unknown but tactile information is available.

The primary contribution of this paper is an algorithm for synthesis of a control policy, utilizing state and force feedback, which is provably stabilizing even during contact mode transitions for systems with possibly non-unique solutions.
To achieve this, we choose a structure for controller and Lyapunov function designed specifically to leverage the complementarity structure of contact.
While verification can be posed as a convex optimization problem, control synthesis is inherently harder. 
This problem is formulated and solved as a bilinear matrix inequality (BMI).

A preliminary version of this paper was presented at the
International Conference on Robotics and Automation
\cite{aydinoglu2019contact}. In this work, extensions are as follows.
\begin{enumerate}
	\item The results are extended to a significantly broader class of systems. 
	\begin{itemize}
		\item The P-matrix assumption (Section \ref{section:background}) is removed which enables design for systems with non-unique solutions (Section \ref{section:LCS_tactile}).
		\item Models where there is a coupling between the contact force and the control loop, including friction models are discussed (Section \ref{section:LCS_tactile}).
		\item Stability analysis (Section \ref{section:stabilization_LCS}, Theorem \ref{theorem_stability_nonunique}) for this broader class of systems is presented.
	\end{itemize}
	\item Better approximations for the sets used in S-procedure (Section \ref{section:cont_design}, \eqref{eq:sol_graph_derivative_2}) are introduced.
	\item A polynomial optimization program (Section \ref{section:cont_design}, \eqref{eq:find_W_poly}) that can describe the non-unique solution sets of linear complementarity problems is introduced.
	\item Four new examples are presented. Three of them are quasi-static friction models with non-unique contact forces. The fourth is a high dimensional example with eight states and ten contacts.
	\item \clr{Results are verified on an experimental setup, and the effectiveness of the proposed method is shown on the task of stabilizing a cart-pole with soft walls.}
\end{enumerate}

\section{Background}
\label{section:background}
We first introduce the definitions and notation used throughout this work. For a positive integer $l$, $\bar{l}$ denotes the set $\{ 1, 2, \ldots, l \}$. Given a matrix $M \in \mathbb{R}^{k \times l}$ and two subsets $I \subseteq \bar{k}$ and $J \subseteq \bar{l}$, we define $M_{I J} = (m_{i j})_{i \in I, j \in J}$. \rev{ For two vectors $a \in \mathbb{R}^m$ and $b \in \mathbb{R}^m$, the notation $0 \leq a \perp b \geq 0$ is used to denote that $a \geq 0, \; b \geq 0, \; a^T b = 0$. } \clr{ The collection of all absolutely continuous functions on a closed interval $[\alpha, \beta]$ is denoted as $AC([\alpha, \beta])$.} 
\rev{The indeterminates are denoted with bold vectors, e.g., $\boldsymbol{x}$.}

\subsection{Linear Complementarity Systems}

A standard approach to modeling robotic systems is through the framework of rigid-body systems with contacts. The continuous time dynamics can be modeled by manipulator equations
\begin{equation}
\label{eq:manipulator}
    M(q) \dot{v} + C(q,v) = Bu + J(q)^T \lambda,
\end{equation}
\clr{ where $q$ represents the generalized coordinates, $v$ represents the generalized velocities, $\lambda \in \mathbb{R}^m$ represents the contact forces, $M(q)$ is the inertia matrix, $C(q,v)$ represents the combined Coriolis, centrifugal and gravitational terms}, $B$ maps the control inputs $u \in \mathbb{R}^k$ into joint coordinates and $J(q)$ is the projection matrix (typically the contact Jacobian). 

The model \eqref{eq:manipulator} is a hybrid dynamical system \cite{alur2000discrete}, \cite{branicky1998unified} where the number of modes scales exponentially with $m$ which arises from distinct combinations of contacts. 
One approach to contact dynamics describes the forces using the complementarity framework where the \clr{ generalized coordinates $q$, velocities $v$ and contact forces $\lambda$ satisfy a set of complementarity constraints:}
\clr{
\begin{equation}
\label{eq:complementarity_constraints}
    \lambda \geq 0, \; \phi(q, v, \lambda) \geq 0, \; \phi(q, v, \lambda)^T \lambda = 0,
\end{equation}
where the function $\phi:\mathbb{R}^p \times \mathbb{R}^m \rightarrow \mathbb{R}^m $ relates the position and velocity of the robot with contact force (\cite{halm2019quasi}, \cite{brogliato1999nonsmooth}, \cite{stewart2000rigid}, \cite{leine2007stability} for more details). 
}
\clr{The complementarity framework is widespread within the robotics community and has been commonly used to simulate contact dynamics \cite{anitescu1997formulating, stewart2000implicit}, leveraged in trajectory optimization \cite{posa2014direct}, stability analysis \cite{haas2016distinction, brogliato2007dissipative}, adaptive control \cite{zavala2001direct}, passivity-based control \cite{moruarescu2010passivity, ccamlibel2002linear}, observer design \cite{menini2002velocity}, trajectory tracking \cite{moruarescu2010trajectory, menini2001asymptotic} and feedback control \cite{brogliato2018feedback, brogliato2000control, brogliato1997control}
of rigid-body systems with contacts. }

The local behavior of \eqref{eq:manipulator} with the constraints \eqref{eq:complementarity_constraints} can be captured by linear complementarity systems \cite{heemels2000linear}, \cite{mayne2001control}.
A linear complementarity system is characterized by: $\bar{A} \in \mathbb{R}^{n_x \times n_x}$, $B \in \mathbb{R}^{n_x \times n_k}$, $\bar{D} \in \mathbb{R}^{n_x \times m}$, $a \in \mathbb{R}^{n_x}$, $\bar{E} \in \mathbb{R}^{m \times n_x}$, $\bar{F} \in \mathbb{R}^{m \times m}$, $H \in \mathbb{R}^{m \times n_k}$, $c \in \mathbb{R}^{m}$ in the following way:
\begin{definition}(Linear Complementarity System)
\label{definition_lcs}
A linear complementarity system (LCS) describes the evolution of two time-dependent trajectories  $\bar{x}(t) \in \mathbb{R}^{n_x}$ and $\lambda(t) \in \mathbb{R}^{m}$ for a given $u(t) \in \mathbb{R}^{n_k}$ and \clr{$\bar{x}(0)$} such that
\begin{equation}
\label{eq:LCS_or}
  \begin{aligned}
    & \qquad \dot{\bar{x}} = \bar{A}\bar{x} + Bu + \bar{D} \lambda + a,\\
    & 0 \leq \lambda \perp \bar{E} \bar{x} +  \bar{F} \lambda + Hu + c \geq 0, \\
  \end{aligned}
\end{equation}
where $\bar{A}$ determines the autonomous dynamics of the state vector $\bar{x}$, $B$ models the effect of the input on the state, $\bar{D}$ describes the effect of the contact forces on the state and $a$ models the constant forces acting on the state.
\end{definition}
The matrices $\bar{E}, \bar{F}, H$\footnote{\clr{ Even though the contact force $\lambda$ does not depend on the input $u$ in \eqref{eq:complementarity_constraints}, local approximations of \eqref{eq:manipulator} and \eqref{eq:complementarity_constraints} can lead to models where the contact force depends on the input under the quasi-static assumption, e.g. \cite{halm2019quasi}, when $v$ depends on the input $u$ and $x = q$. An important example where the contact force depends on the input is quasi-static friction models (Sections \ref{sub:friction}, \ref{sub:friction2}, \ref{sub:friction3}) and this affect is captured by the $Hu$ term in the LCS.}} 
and the vector $c$ capture the relationship between the contact force $\lambda$, the state vector $\bar{x}$ and the input $u$.
Note that the contact forces $\lambda$ are always non-negative which holds for basic model of normal force and slack variables are typically used to represent sign-indefinite frictional forces.
\eqref{eq:LCS_or} implies that either $\lambda = 0$ or $\bar{E}\bar{x} + \bar{F} \lambda ~ + ~ Hu ~ + ~ c ~ = ~ 0$, encoding the multi-modal dynamics of contact.
Due to this complementarity structure, an LCS is a compact representation, as the variables and constraints scale linearly with $m$, rather than with the potential $2^m$ hybrid modes \cite{camlibel2001complementarity}, \cite{lin2009stability}.

\subsection{Linear Complementarity Problem}
A linear complementarity system is an ordinary differential equation (ODE) coupled with a variable that is the solution of a linear complementarity problem.
\rev{ Since linear complementarity problems play an important role in understanding and analyzing the LCS, some definitions and results from the theory of linear complementarity problems are summarized \cite{cottle2009linear}. }
\begin{definition}(Linear Complementarity Problem)
	Given ${F \in \mathbb{R}^{m \times m}}$ and a vector $w \in \mathbb{R}^m$, the linear complementarity problem $\text{LCP}(w,F)$  is the following mathematical program:
	\begin{alignat}{2}
	\label{eq:LCP}
	\notag & \underset{}{\text{find}} && \lambda \in \mathbb{R}^m \\
	& \text{subject to}  \quad && 0 \leq \lambda \perp F \lambda + w \geq 0.
	\end{alignat}
\end{definition}
\rev{ For a given $F$ and $w$, the LCP may have multiple solutions or none at all. Hence, the solution set of the linear complementarity problem $\text{LCP}(w,F)$ is
}
\begin{equation*}
\text{SOL}(w,F) = \{ \lambda : 0 \leq \lambda \perp F \lambda + w \geq 0 \}.
\end{equation*}
In this work, we will consider LCP's where $\text{SOL}(w,F)$ can have more than one element for a given $F$ and $w$. As a special case of this, we mention a particular class of LCP's that are guaranteed to have unique solutions.

\begin{definition}(P-Matrix)
	A matrix $F \in \mathbb{R}^{m \times m}$ is a P-matrix, if the determinants of all of its principal sub-matrices are positive; that is, $\text{det}(F_{\alpha \alpha}) > 0$ for all $\alpha \subseteq \{ 1, \ldots, m \} $.
\end{definition}
If $F$ is a P-matrix, then the solution set $\text{SOL}(w,F)$ is a singleton for any $w \in \mathbb{R}^m$ \cite{shen2005linear}. 
\rev{ If the unique element of $\text{SOL}(w,F)$ is $\psi(w)$, then it is a piecewise linear function in $w \in \mathbb{R}^m$, hence is Lipschitz continuous and directionally differentiable.}

If $\bar{F}$ is a P-matrix, one can represent an LCS in a more compact manner.
The linear complementarity system in \eqref{eq:LCS_or} is equivalent to the dynamical system
\begin{align}
\label{eq:system_rep}
&\dot{\bar{x}} = \bar{A} \bar{x} + Bu + \bar{D} \lambda(\bar{x},u),
\end{align}
where $\lambda(\bar{x},u)$ corresponds to the unique element of $\text{SOL}(\bar{E}\bar{x}+Hu+c,\bar{F})$ for every state vector $\bar{x}$. Notice that \eqref{eq:system_rep} is only a an alternative representation of \eqref{eq:LCS_or} and still has the same structure as the LCS.

\subsection{Sum-of-squares}
In this work (Section \ref{section:cont_design}, \eqref{eq:find_W_poly}), describing the non-unique solution sets of LCP's is posed as a question of non-negativity of polynomials on basic semialgebraic sets. \rev{ Towards this direction, sum-of-squares (SOS) optimization is used. }

A multivariate polynomial $p(x)$ is a sum-of-squares (SOS) if there exist polynomials $q_i(x)$ such that
\begin{equation*}
	p(x) = \sum_i q_i^2(x).
\end{equation*}
The existence of a sum-of-squares decomposition of a polynomial can be decided by solving a semidefinite programming feasibility problem \cite{parrilo2003semidefinite}, which is a convex optimization problem.
We represent the semialgebraic conditions using the S-procedure technique \cite{stengle1974nullstellensatz}, \cite{boyd1994linear}. For example, to show that \cite{posa2017balancing}
\begin{equation*}
	f(x) \geq 0, \quad \forall x \in \{z: g(z) \geq 0, h(z ) = 0    \},
\end{equation*}	
it is sufficient to find polynomials $\sigma_1(x), \sigma_2(x), q(x) $ s.t.
\begin{align}
	\label{eq:SOS}
	\notag &\sigma_1(x) f(x) - \sigma_2(x) g(x) - q(x) h(x) \geq 0, \\
	&\sigma_1(x) - 1 \geq 0, \\
	\notag &\sigma_2(x) \geq 0.
\end{align}
If constraints are in the form of \eqref{eq:SOS} and the objective function is linear in the coefficients of any unknown/free polynomials, then the optimization problem can be represented as a semidefinite program (SDP) using the SOS relaxation.

\section{Linear Complementarity Systems with Tactile Feedback}
\label{section:LCS_tactile}
In this section, we present a tactile feedback controller where the input is dependent both on the state and the contact force $\big( u = u(x,\lambda) \big)$,  unlike the common approach of designing controllers only using the state feedback, $ \big( {u = u(x)} \big)$. 
\rev{ The section concludes with a description of complementarity models with such tactile feedback controllers.}

\subsection{Tactile Feedback and Related Complementarity Models}
\label{subsection:feedback}
We introduce the tactile feedback controller:
\begin{equation}
	\label{eq: controller}
	u(\bar{x},\lambda)~=~K\bar{x} + L \lambda,
\end{equation}
where $K \in \mathbb{R}^{n_k \times n_x}$ and $L \in \mathbb{R}^{n_k \times m}$. \rev{ Using this control law, \eqref{eq:LCS_or} can be transformed into the following LCS:}
\begin{equation}
\label{eq:LCS}
\begin{aligned}
& \qquad \dot{x} = Ax + D\lambda + a,\\
& 0 \leq \lambda \perp Ex +  F \lambda + c \geq 0,
\end{aligned}
\end{equation}
where $A \in \mathbb{R}^{n \times n}$, $D \in \mathbb{R}^{n \times m}$, $a \in \mathbb{R}^{n}$, $E \in \mathbb{R}^{m \times n}$, $F \in \mathbb{R}^{m \times m}$, $c \in \mathbb{R}^{m}$.

\clr{ If the contact force does not depend on the input ($H=0$), then application of the control law \eqref{eq: controller} trivially produces \eqref{eq:LCS} with $A~=~ \bar{A}+BK$, $D = \bar{D}+BL$, $E = \bar{E}$, and $F = \bar{F}$. In this case, note that $x = \bar{x}$ and $n = n_x$.}

\rev{ Next, consider the case where the contact force depends on the input ($H \neq 0$).} Since the input $u = u(\bar{x},\lambda)$ similarly depends on the contact force, this introduces an algebraic loop. One might attempt to resolve this loop by simultaneously solving for both $u$ and $\lambda$, leading to the closed-loop LCS:
\begin{equation*}
\begin{aligned}
& \dot{x} = (\bar{A}+BK)x + (\bar{D}+BL)\lambda + a,\\
& \qquad 0 \leq \lambda \perp E x +  F \lambda + c \geq 0,
\end{aligned}
\end{equation*}
\clr{ where $x = \bar{x}$, $E = \bar{E} + HK$ and $F = \bar{F} + HL$.} Observe that the matrix $F$ depends on the choice of the contact gain matrix $L$. Due to this dependency, the cardinality of the solution set $\text{SOL}(Ex + c,F)$ for a given $x$ might change depending on the value of $L$. \rev{ This is illustrated via an example.}
\begin{example}
\label{example:example}
Consider the complementarity constraint:
\begin{equation*}
	0 \leq \lambda \perp x + u + \lambda \geq 0,
\end{equation*}
where $x, u, \lambda \in \mathbb{R}$.
If $u$ is independent of $\lambda$, observe that $\text{SOL}(x+u, F)$ is a singleton for all pairs $(x,u)$ since $F = [1]$ is a P-matrix. In this case, the contact force $\lambda_o(x,u)$ is equal to
\begin{equation*}
	\lambda_o(x,u) = \max \{ 0, -x-u \}.
\end{equation*}
However, for some choices of force-dependent inputs, this is no longer the case. From $u = L \lambda$, it follows that $F = [1 + L]$. For the case $L=-1$, the LCP for the closed-loop system is
\begin{equation*}
0 \leq \lambda \perp x \geq 0.
\end{equation*}
The solution set is then:
\begin{equation*}
	\text{SOL}(x,F=0) = \begin{cases}
	\{0 \} &  \mbox{if} \quad x > 0, \\
	[0, \infty) & \mbox{if} \quad x = 0, \\
	\emptyset & \mbox{if} \quad x < 0.
	\end{cases}
\end{equation*}
\rev{ There are infinitely many solutions for $x = 0$ and no solutions for $x < 0$.}
\end{example}

Furthermore, resolving the algebraic loop by solving simultaneously for the contact force and the input is not physically realistic since control policies can not instantaneously respond to tactile measurements. As illustrated in Example \ref{example:example}, it is also mathematically problematic. 
Therefore, we will use the standard approach of modeling delay.
\rev{ Specifically, the following low-pass filter model captures the input delay: }
\begin{equation}
\label{eq:low_pass_filter}
\dot{\tau} = \kappa (u - \tau),
\end{equation}
where $\kappa \in \mathbb{R}^+$ is the rate parameter. Using the low-pass filter model, we obtain the linear complementarity system:
\begin{equation}
\begin{aligned}
\label{eq:LCS_low_pass}
& \quad \dot{ \bar{x} } = \bar{A} \bar{x} + B \tau + \bar{D} \lambda + a,\\
& \qquad \quad \dot{\tau} = \kappa (u - \tau),\\
& 0 \leq \lambda \perp \bar{E} \bar{x} +  F \lambda + H \tau + c \geq 0,
\end{aligned}
\end{equation}
Observe that the LCS model in \eqref{eq:LCS_low_pass} has the same form with \eqref{eq:LCS} with the input \eqref{eq: controller}:
\begin{equation*}
\begin{aligned}
& \dot{x} = \begin{bmatrix}
\bar{A} & B \\
\kappa K & - \kappa I
\end{bmatrix} x  + , \begin{bmatrix}
\bar{D} \\ \kappa L
\end{bmatrix} \lambda + \begin{bmatrix}
a \\ 0
\end{bmatrix}, \\ 
& 0 \leq \lambda \perp \begin{bmatrix}
\bar{E} & H
\end{bmatrix} x +  F \lambda  + c \geq 0,
\end{aligned}
\end{equation*}
where $x = \begin{bmatrix} \bar{x}^T & \tau^T \end{bmatrix}^T$. Observe that the delay decouples $u$ and $\lambda$ so the matrix $F$ does not depend on the contact gain matrix $L$. 
\rev{Notice that the state is augmented ($n > n_x$) to obtain \eqref{eq:LCS}.}
\rev{ Alternatively, one could add delay to the sensor dynamics: }
\begin{equation*}
\dot{\tau}_s = \kappa_s (\lambda - \tau_s).
\end{equation*}
While this approach would similarly resolve the algebraic loop, in this work we found out that modeling input delay produced better numerical results when combined with the algorithmic approach in Section \ref{section:cont_design}.

Using the control format in \eqref{eq: controller}, for notational compactness, we will now exclusively consider closed-loop LCS in the form of \eqref{eq:LCS}. 
\rev{As a result of filtering, the matrix $F$ will be independent of the tactile feedback gain $L$.}

\subsection{Solution Concept}
We introduce a solution concept for complementarity systems \eqref{eq:manipulator} and \eqref{eq:LCS} similar to (\cite{ccamlibel2002solution}, Definition 3.6).
\begin{definition}
	\label{solution}
	A pair of functions $(x(t), \lambda(t))$ is a solution of the complementarity system,
	\begin{equation*}
	\begin{aligned}
	& \quad \dot{x} = f(x,\lambda),\\
	& 0 \leq \lambda \perp \Phi(x,\lambda) \geq 0,
	\end{aligned}
	\end{equation*}
	where $f:\mathbb{R}^n \times \mathbb{R}^m \rightarrow \mathbb{R}^n$ and $\Phi:\mathbb{R}^n \times \mathbb{R}^m \rightarrow \mathbb{R}^m$
	with the initial condition $x(0) = x_0$ if:
	\begin{equation*}
	\begin{aligned}	
	& x(t) \in AC([0, T]), \; \forall \; T \geq 0, \\
	& \dot{x}(t) = f(x(t),\lambda(t)) \; \text{for almost all} \; t \in \mathbb{R}^+,\\
	& 0 \leq \lambda(t) \perp \Phi(x(t), \lambda(t) ) \geq 0 \; \text{for almost all} \; t \in \mathbb{R}^+,\\
	& \lambda(t) \; \text{is almost everywhere differentiable.}
	\end{aligned}
	\end{equation*}
\end{definition}

It is important to note that we restrict ourselves to complementarity systems where the state, $x(t)$, is absolutely continuous which is well-studied in the literature (\cite{shen2007semicopositive,brogliato2010existence,ccamlibel2002linear}). 
The models considered in this work have no jumps (e.g. impact). \rev{ Observe that $\lambda(t)$ can be discontinuous and it is assumed that $\lambda(t)$ is almost everywhere differentiable moving forward.} \rev{Consider the following proposition by Camlibel et al. \cite{camlibel2006lyapunov}:}
\begin{proposition} 
	\label{existence_uniqueness}
	For every $x_0$, the LCS \eqref{eq:LCS} has a unique 
	\rev{$C^1$ (hence absolutely continuous)} trajectory $x(t)$ defined for all $t \geq 0$ if and only if the set $D \text{SOL}(Ex+c,F)$ is a singleton for every $x \in \mathbb{R}^n$.
\end{proposition}

\rev{
Throughout this work, focus is on LCS models where $\bar{D} \text{SOL}(Ex+c,F)$ is a singleton. Note that unlike $x(t)$, the trajectory $\lambda(t)$ is not necessarily unique which is observed in friction models (Section \ref{section:examples}). 
Recent results indicate that it may, ultimately, be possible to eliminate this assumption on $\bar{D} \text{SOL}(Ex+c,F)$ \cite{brogliato2010existence, brogliato2020dynamical}, though such exploration is outside the scope of this work.
}

\rev{
In Sections \ref{section:stabilization_LCS} and \ref{section:cont_design}, this structure is leveraged to similarly ensure that $L \text{SOL}(Ex+c,F)$ is a singleton for all $x$.
Since we consider models such that $\bar{D} \text{SOL}(Ex+c,F)$ is a singleton and $F$ is independent of the controller (Section \ref{section:LCS_tactile}), 
the condition that $L \text{SOL}(Ex+c,F)$ is a singleton suffices to ensure that $D \text{SOL}(Ex+c,F)$ is also a singleton. The trajectories of the closed-loop system \eqref{eq:LCS} remain absolutely continuous as long as $L \text{SOL}(Ex+c,F)$ is a singleton for all $x$ (following Proposition 6).
}

\rev{Moving forward, denote $\mathcal{S}(t_0, x_0)$ as the set of all trajectories $x(t)$, with $t \geq t_0$, such that $x(0) = x_0$. The dependency on the LCS parameters is suppressed for ease of notation. Observe that $S(t_0, x_0)$ is also a singleton following Proposition \ref{existence_uniqueness} if $D \text{SOL}(Ex+c,F)$ is a singleton for every $x$.
}

\section{Stabilization of the Linear Complementarity System}
\label{section:stabilization_LCS}
\rev{In this section, conditions for stabilization using non-smooth monontonic Lyapunov functions and contact-aware controllers are constructed.}

\rev{Notions of stability from \cite{smirnov2002introduction} are adopted}. If $F$ is a P-matrix, these are equivalent to the notions of stability for differential equations where the right-hand side is Lipschitz continuous, though possibly non-smooth \cite{camlibel2006lyapunov}, \cite{khalil2002nonlinear}.
\begin{definition}
	The equilibrium $x_e$ of LCS \eqref{eq:LCS} is
	\begin{IEEEenumerate}
		\item stable in the sense of Lyapunov if, given any $\epsilon > 0$, there exists $\delta > 0$ such that
		\begin{equation*}
			|| x_e - x_0 || < \delta \implies ||x(t) - x_e || < \epsilon \; \forall t \geq 0
		\end{equation*}
		for any $x_0$ and $x(t) \in \mathcal{S}(0, x_0)$.
		\item asymptotically stable if it is stable and $\delta > 0$ exists s.t.
		\begin{equation*}
		|| x_e - x_0 || < \delta \implies \lim_{t \rightarrow \infty} x(t) = x_e
		\end{equation*}
		for any $x_0$ and $x(t) \in \mathcal{S}(0, x_0)$.
	\end{IEEEenumerate}
\end{definition}

\subsection{Non-Smooth Lyapunov Function}

In Lyapunov based analysis and synthesis methods, one desires to search over a wide class of functions. \rev{ Here, piecewise quadratic Lyapunov functions are considered}. They are more expressive than a Lyapunov function common to all modes (as was used in \cite{posa2015stability}), which makes it a more powerful choice than a single quadratic Lyapunov function \cite{johansson1997computation}.
\rev { Towards this direction, consider a variant of the Lyapunov function introduced in \cite{camlibel2006lyapunov}: }
\begin{equation}
\label{eq:lyap_func}
    V(x,\lambda) = x^T P x + 2 x^T Q \lambda + \lambda^T R \lambda + p^T x + r^T \lambda + z,
\end{equation}
where $P \in \mathbb{R}^{n \times n}$, $Q \in \mathbb{R}^{n \times m}$, $R \in \mathbb{R}^{m \times m}$, $p \in \mathbb{R}^n$, $r \in \mathbb{R}^m$, and $z \in \mathbb{R}$. 
The Lyapunov function \eqref{eq:lyap_func} is quadratic in terms of the pair $(x,\lambda)$. If $F$ is a P-matrix, it is piecewise quadratic in $x$ since $\lambda = \lambda(x)$ is a piecewise affine function.
For example, if all contact forces are inactive, $\lambda = 0$, then $V(x,\lambda(x)) = x^T P x + p^T x + z$.
Even though $V$ is non-smooth, it is locally Lipschitz continuous with respect to $x$ if $F$ is a P-matrix \cite{camlibel2006lyapunov}.

\begin{figure}[t]
	\hspace{-0.5cm}
	\includegraphics[width=1.1\columnwidth]{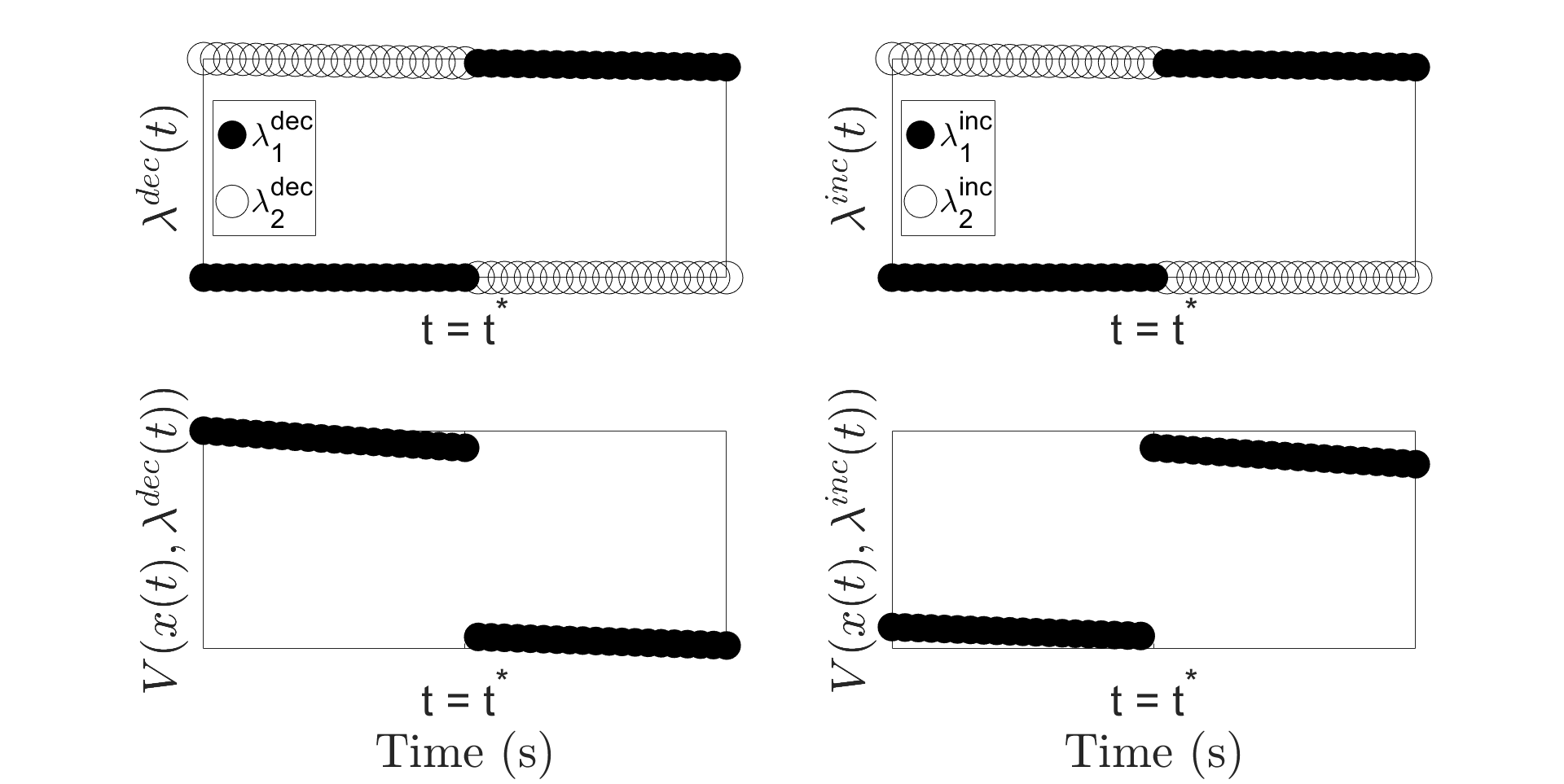}
	\caption{ \clr{Two different solutions for the Lyapunov function. For the solution $\lambda^\text{dec}$, $\lambda_1^\text{dec}$ and $\lambda_2^\text{dec}$ represent the first and second elements of the solution vector respectively (similarly for $\lambda^\text{inc}$).} }
	\label{example_comp}
\end{figure}

\clr{ If $\text{SOL}(Ex+c,F)$ is not a singleton then $\lambda(t)$ can be discontinuous in $t$ due to the multi-valued nature of $\text{SOL}(Ex(t)+c,F)$. Similarly, $V(x(t), \lambda(t))$ can be discontinuous due to the terms $Q \text{SOL}(Ex(t)+c,F)$, $\text{SOL}(Ex(t)+c,F)^T R \text{SOL}(Ex(t)+c,F)$ and $r^T \text{SOL}(Ex(t)+c,F)$.  \rev{ Next, it is shown that without appropriate restrictions, these discontinuities imply that such functions $V$ cannot be valid \rev{monotonic} Lyapunov functions. } } 
\begin{example}
	\label{eg:lyap_jump}
	Consider the LCS:
	\begin{equation*}
	\begin{aligned}
	& \qquad \dot{x} = -x + \lambda_1 + \lambda_2, \\
	& 0 \leq \lambda_1 \perp x + \lambda_1 + \lambda_2 \geq 0,\\
	&  0 \leq \lambda_2 \perp x + \lambda_1 + \lambda_2 \geq 0,
	\end{aligned}
\end{equation*}
where $x, \lambda_1, \lambda_2 \in \mathbb{R}$. 
As seen in Figure \ref{example_comp}, the Lyapunov function jumps at $t=t^*$ for a solution $(x(t), \lambda^{\text{dec}}(t))$ and the Lyapunov function decreases ($V(t^*_+) > V(t^*_-)$).
Then, consider $\lambda^{inc}(t)$ that jumps at $t = t^*$ where $\lambda^{inc}_- = \lambda^{dec}_+$ and $\lambda^{inc}_+ = \lambda^{dec}_-$. The Lyapunov functions value increases after the jump at $t=t^*$ for the solution $(x(t), \lambda^{inc}(t))$ by construction as seen in Figure \ref{example_comp}.
\end{example}

If the function $V$ jumps at a time $t^*$ for one solution and its value decreases, then another solution exists where $V$ jumps at $t^*$ and the function's value increases as illustrated in Example \ref{eg:lyap_jump}.
Hence, the Lyapunov function cannot decrease monotonically along all solutions due to the multi-valued nature of $\text{SOL}(Ex + c, F)$. \clr{ For this reason, we focus on mappings $W$ such that $W \text{SOL}(Ex+c,F)$ is single-valued.
Using such mappings, one can parameterize $Q$, $R$ and $r$ such that $Q \text{SOL}(Ex(t)+c,F)$, $\text{SOL}(Ex(t)+c,F)^T R \text{SOL}(Ex(t)+c,F)$ and $r^T \text{SOL}(Ex(t)+c,F)$ are single-valued for all $t$ even when $F$ is not a P-matrix. 
} 
\begin{proposition}(\cite{camlibel2006lyapunov}, Proposition 3.9)
	\label{proposition_W}
	Assume that $W \text{SOL}(q,F)$ is a singleton for all $q$ where  $W \in \mathbb{R}^{n_w \times m}$. Then, the map $q \mapsto W \text{SOL}(q,F)$ is a continuous piecewise linear function of $q$.
\end{proposition}
One can construct a Lyapunov function using a matrix $W$ as in Proposition \ref{proposition_W} where $Q \text{SOL}(Ex+c)$, $R \text{SOL}(Ex+c)$, and $r^T \text{SOL}(Ex+c)$ are singletons for all $x \in \mathbb{R}^n$ as follows:
\begin{align}
\label{eq:lyap_nonunique}
V(x, \lambda) = x&^T P x + 2 x^T \tilde{Q} W \lambda + \lambda^T  W^T \tilde{R} W \lambda \\
& + p^T x + \tilde{r}^T W \lambda + z, \nonumber
\end{align}
\clr{ where $W \in \mathbb{R}^{n_w \times m}$, $\tilde{Q} \in \mathbb{R}^{n \times n_w}$, $\tilde{R} \in \mathbb{R}^{n_w \times n_w}$, $\tilde{r} \in \mathbb{R}^{n_w}$ and $W \text{SOL}(Ex+c,F)$ is a singleton. }
\rev{ Next, it is shown that $V$ is a non-smooth, continuous piecewise quadratic function in $x$ and is locally Lipschitz continuous which are helpful properties in establishing stability results.}
\begin{lemma}
	\label{lyap_continuity_nonp}
	The Lyapunov function\footnote{$\lambda(x)$ is the set-valued function $\lambda(x) = \text{SOL(Ex+c,F)}$.} $V(x,\lambda(x) )$ as in \eqref{eq:lyap_nonunique} is locally Lispchitz continuous in $x$. Furthermore, $\bar{V}(t) = V(x(t), \lambda(t)) \in AC([0, T])$ for all $T \geq 0$ for the solutions as in Definition \ref{solution}.
\end{lemma}
\begin{IEEEproof}
	Since $W \lambda(x)$ is Lipschitz continuous in $x$, $V(x,\lambda(x) )$ is locally Lispchitz continuous in $x$. Because $x(t)$ is absolutely continuous and $V(x, \lambda(x))$ is locally Lipschitz continuous, $V$ is absolutely continuous in time.
\end{IEEEproof}
\rev{ From this point onward, without loss of generality, the Lyapunov function as defined in \eqref{eq:lyap_nonunique} is used.  Observe that if $F$ is a P-matrix, one can trivially choose $W=I$. 
For many practical examples, it is possible to find such a matrix $W$. However, such a $W$ is not guaranteed to exist when $F$ is not a P-matrix.	
In Section \ref{section:cont_design}-C, it is shown how to generate $W$ algorithmically. }

\begin{remark}
	Similar to the Lyapunov function, the input \eqref{eq: controller} is not necessarily continuous in time. If one desires a controller that is continuous in time, then the parametrization
	\begin{equation}
		\label{eq:cont_used}
	u(\bar{x},\lambda) = K \bar{x} + \tilde{L} W \lambda,
	\end{equation}
	leads to a controller $u$ that is continuous in time. 
	\rev{As discussed in Section \ref{section:examples}-B, it is required that $D \text{SOL}(Ex+c,F)$ be a singleton for all $x$. Therefore, we restrict the controller to be of the form \eqref{eq:cont_used}.
	For all of the examples in Section \ref{section:examples}, the parametrization $L = \tilde{L} W$ is used. }
\end{remark}

\subsection{Conditions for Stabilization}
\rev{ Now, conditions for stability in the sense of Lyapunov are constructed with the controller gains $K$ and $L$ as in \eqref{eq: controller}, and the piecewise quadratic Lyapunov function $V$. These conditions will be the building blocks for the controller design method proposed in Section \ref{section:cont_design}. }
\begin{theorem}
	\label{theorem_stability_nonunique}
	Consider the linear complementarity system \eqref{eq:LCS}, and the Lyapunov function \eqref{eq:lyap_nonunique} with $W$ such that $W \text{SOL} (Ex+c,F)$ is a singleton for all $x$. Assume there exists a solution for every $x_0$ and $x_e = 0$ is an equilibrium.
	If for all solutions $(x(t), \lambda(t))$\footnote{Dependence on $x_0$ and LCS parameters is suppressed.}, there exist strictly positive constants $\gamma_1$, $\gamma_2$, matrices $K,L$\footnote{$\frac{d \bar{V}(t)}{d t}$ depends on $K$ and $L$ since $\dot{ x}$ is a function of $K$ and $L$.} and a function $V$ such that
	\begin{align*}
	\gamma_1 ||x(t)||_2^2 \leq V(x(t), \lambda(t)) \leq \gamma_2 ||x(t)||_2^2,
	\end{align*}
	and $\frac{d \bar{V}(t)}{d t} \leq 0$ for almost all $t$, then $x_e = 0$ is Lyapunov stable. Furthermore, if there exists a strictly positive constant $\gamma_3$ such that $\frac{d \bar{V}(t)}{d t} \leq - \gamma_3 ||x(t)||_2^2$ for almost all $t$, then $x_e = 0$ is exponentially stable.
\end{theorem}
\begin{IEEEproof}
	Let the solution $(x(t), \lambda(t))$ be arbitrary. Following Lemma \ref{lyap_continuity_nonp}, $\bar{V}(t)$ is absolutely continuous and almost everywhere differentiable on $ [ 0, T ]$ for all $T$. Then we have
	\begin{equation*}
	\bar{V}(t) = \bar{V}(0) + \int_0^t \dot{\bar{V}}(s) ds \leq \bar{V}(0),
	\end{equation*}
	since $\dot{\bar{V}} \leq 0$ for almost all $t \in \mathbb{R}^+$. Since $V$ is bounded and non-increasing, the rest follows from standard arguments for Lyapunov stability.
	
	In order to prove exponential stability, observe that $\dot{\bar{V}}(t)~\leq~-\gamma_3 ||x(t)||_2^2$. Hence, it follows that
	\begin{equation*}
	||x(t)||_2^2 \leq \frac{1}{\gamma_1}\bar{V}(0) - \frac{\gamma_3}{\gamma_1} \int_0^t ||x(s)||_2^2 ds.
	\end{equation*}
	Using \clr{Gr\"onwall's} inequality, it follows that
	\begin{equation*}
	||x(t)||_2^2 \leq \frac{1}{\gamma_1} \bar{V}(0) e^{-\frac{\gamma_3}{\gamma_1} t } \leq \frac{\gamma_2}{\gamma_1} ||x_0||_2^2 e^{-\frac{\gamma_3}{\gamma_1} t }.
	\end{equation*}
	Hence we conclude that the equilibrium is exponentially stable.
\end{IEEEproof}

\rev{ Theorem \ref{theorem_stability_nonunique} establishes sufficient conditions to stabilize the LCS in \eqref{eq:LCS}.}
\rev{ In Section \ref{section:cont_design}, it is shown how Theorem \ref{theorem_stability_nonunique} can be used to algorithmically synthesize a controller. }
\rev{ Next, observe that an upper-bound always exists under certain assumptions. }
\begin{remark}
	If $c \geq 0$ and $z = 0$, there exists a $\gamma_2$ such that $V(x, \lambda) \leq \gamma_2 ||x||_2^2$ since $W \lambda(x) \leq \rho ||x||_2$ for all $x$ for some $\rho$.
\end{remark}

For this special case, an upper-bound always exists and one does not need to verify that $V$ is upper-bounded when algorithmically synthesizing a controller (Section \ref{section:cont_design}).

\section{Controller Design as a Bilinear Matrix Inequality Feasibility Problem}
\label{section:cont_design}

\clr{Controller design for complementarity systems in the form of BMI's have been explored before for the case where $F$ in \eqref{eq:LCS} is zero and without tactile feedback ($L=0$) \cite{brogliato2007dissipative}. 
\rev{ In this section, these results are extended and it is shown how Theorem \ref{theorem_stability_nonunique} can be used to algorithmically synthesize a controller when there is tactile feedback and the matrix $F$ is non-zero. 
Then, the controller design problem is converted into a bilinear matrix inequality (BMI) feasibility problem. A convex optimization program is proposed to find a matrix $W$ such that $W \text{SOL}(Ex + c, F)$ is a singleton for all $x$.  } }

\subsection{Sufficient Conditions for Stabilization}

\clr{ The sufficient conditions in Theorem \ref{theorem_stability_nonunique} need to be satisfied for all solutions of the LCS \eqref{eq:LCS}.} Now, we will transform them into matrix inequalities over two basic semialgebraic sets $\Gamma_{\text{SOL}}(E,F,c)$ and $\Gamma'_{\text{SOL}}(E,F,c)$. \rev{ Define the set: }
\begin{equation*}
	\Gamma_{\text{SOL}}(E,F,c) = \{ (\boldsymbol{x},\boldsymbol{\lambda}) : 0 \leq \boldsymbol{\lambda} \perp E \boldsymbol{x} + c + F \boldsymbol{\lambda} \geq 0 \},
\end{equation*}
where $(\boldsymbol{x},\boldsymbol{\lambda)} \in \Gamma_{\text{SOL}}(E,F,c)$ are represented as quadratic inequalities.
\rev{ Similarly, define the following set: }
\begin{align}
\label{eq:sol_graph_derivative_2}
\notag \Gamma'_{\text{SOL}}&(E,F,c) = \{ (\boldsymbol{x},\boldsymbol{\lambda},\boldsymbol{\dot{\lambda}}) \big| \exists \boldsymbol{\rho}, \boldsymbol{\mu} : \boldsymbol{\lambda} \in SOL(E\boldsymbol{x}+c,F), \\
& E \dot{x} + F \boldsymbol{\dot{\lambda}} + \boldsymbol{\rho} = 0, \; \boldsymbol{\lambda}_i \boldsymbol{\rho}_i = 0, \boldsymbol{\dot{\lambda}}_i + \boldsymbol{\mu}_i = 0, \\
\notag & (E_i^T \boldsymbol{x} + F_i^T \boldsymbol{\lambda} + c_i) \boldsymbol{\mu}_i = 0, \boldsymbol{\mu}_i \boldsymbol{\rho}_i = 0  \},
\end{align}
where $\dot{x} = A \boldsymbol{x} + D \boldsymbol{\lambda} + a$ and $\boldsymbol{\mu}, \boldsymbol{\rho}$ are slack variables. Here $\boldsymbol{\dot{\lambda}}$ expresses the time derivative of the force. \rev{ Next, we express the matrix inequalities over semialgebraic sets where a method similar to construction of contact LCP's in (\cite{brogliato1999nonsmooth}, Section 5.1.2.1) is used. }
\clr{
\begin{proposition}
	\label{constraints}
	If the inequalities
	\begin{align}
	\label{eq:ineq1}
	& \gamma_1 ||\boldsymbol{x}||_2^2 \leq V(\boldsymbol{x},\boldsymbol{\lambda}) \leq \gamma_2 ||\boldsymbol{x}||_2^2, \; \forall (\boldsymbol{x},\boldsymbol{\lambda}) \in \Gamma_{\text{SOL}}, \\
	\label{eq:ineq2}
	& \nabla_{\boldsymbol{x}} V(\boldsymbol{x},\boldsymbol{\lambda})^T \dot{x} + \nabla_{\boldsymbol{\lambda}} V(\boldsymbol{x},\boldsymbol{\lambda})^T \boldsymbol{\dot{\lambda}} \leq 0, \; \forall (\boldsymbol{x}, \boldsymbol{\lambda}, \boldsymbol{\dot{\lambda}}) \in \Gamma'_\text{SOL},
	\end{align}
 	hold for the LCS \eqref{eq:LCS} where $\dot{x} = A \boldsymbol{x} + D \boldsymbol{\lambda} + a$,
	then the following inequalities hold for all solutions $(x(t), \lambda(t))$ of the LCS
	\begin{align}
	\label{eq:ineq11}
	& \gamma_1 ||x(t)||_2^2 \leq V(x(t), \lambda(t)) \leq \gamma_2 ||x(t)||_2^2, \\
	\label{eq:ineq22}
	&\frac{d}{dt}V(x(t), \lambda(t)) \leq 0,
	\end{align}
	for almost all $t \geq 0$.
\end{proposition}
}
\begin{IEEEproof}
	Consider an arbitrary solution, $(x(t), \lambda(t) )$ of \eqref{eq:LCS}. First we will show that \eqref{eq:ineq1} implies \eqref{eq:ineq11}.
	From Definition \ref{solution}, it follows that $\lambda(t) \in \text{SOL}(Ex+c,F)$ and $(x(t), \lambda(t)) \in \Gamma_{\text{SOL}}(E,F,c)$ for almost all $t \geq 0$. The result follows from \eqref{eq:ineq1}.
	
	Next, we show that \eqref{eq:ineq2} implies \eqref{eq:ineq22}. We show that
	\begin{align}
		\label{eq:one}
		\lambda(t) \in \text{SOL}(Ex(t) + c, F), \\
		\label{eq:two}
		\lambda_i(t) > 0 \implies E_i^T \dot{x}(t) + F_i^T \dot{\lambda}(t) = 0,  \\
		\label{eq:three}
		E_i^T x(t) + F_i^T \lambda(t) + c_i > 0 \implies \dot{\lambda}_i(t) = 0, \\
		\label{eq:four}
		\left.
		\begin{array}{rr}
		\lambda_i = 0  \\
		E_i^T x + F_i^T \lambda + c = 0
		\end{array}
		\right\} \implies \begin{matrix} \dot{\lambda}_i = 0 \quad  \text{or} \\ E_i^T \dot{x} + F_i^T \dot{\lambda} = 0, \end{matrix}
	\end{align}
	hold for almost all $t\geq 0$ where dependency on $t$ in \eqref{eq:four} is suppressed for space limitations, $\dot{\lambda}(t) = \frac{d \lambda}{dt}$, and \eqref{eq:one} directly follows from the definition of solution.
	
	We define $n_i(t) = E_i^T x(t) + F_i^T \lambda(t) + c$ for notational simplicity.
	To prove \eqref{eq:two}-\eqref{eq:four}, observe that for almost all $t\geq0$, there exists an $\epsilon > 0$ such that both $\lambda_i(t)$ and $n_i(t)$ are continuous in the interval $[t-\epsilon, t+\epsilon]$.
	For almost all $t\geq0$ if $\lambda_i(t) > 0$, then $n_i(t) = 0$ for a neighborhood around $t$ hence
	$E_i^T \dot{x}(t) + F_i^T \dot{\lambda}(t) = 0$ and \eqref{eq:two} follows.
	Similarly observe that if $n_i(t) > 0$, then $\lambda(t) = 0$ and $\dot{\lambda}(t) = 0$ as in \eqref{eq:three}. \eqref{eq:four} follows from the the fact that both $\lambda_i(t)$ and $n_i(t)$ cannot be positive at the same time.
	
	Suppose \eqref{eq:one}-\eqref{eq:four} hold at some time $t^*$ and consider $x_* = x(t^*)$, $\lambda_* = \lambda(t^*)$, $\dot{\lambda}_* = \lambda(t^*)$ and $\dot{x}_* = \dot{x}(t^*)$. We will show that $(x_*, \lambda_*, \dot{\lambda}_*) \in \Gamma'_\text{SOL}(E,F,c)$. 
	
	There are 3 cases. First consider the case where $\lambda_{i,*} > 0$ and therefore $(E_i^T x_{*} + F_i^T \lambda_* + c_i) = 0 $. Observe that all equalities in \eqref{eq:sol_graph_derivative_2} are satisfied with $\rho_{i,*}=0$ and $\mu_{i,*}=-\dot{\lambda}_{i,*}$. 
	
	For the case where $\lambda_{i,*} = 0$ and $(E_i^T x_{*} + F_i^T \lambda_* + c_i) > 0 $, all equalities are satisfied with $\rho_{i,*} = -E_i^T \dot{x}_* - F \dot{\lambda}_{i,*}$ and $\mu_{i,*}=0$. 
	
	For the last case where both $(E_i^T x_{*} + F_i^T \lambda_* + c_i) = \lambda_{i,*} = 0 $, the equalities are satisfied with either $\rho_{i,*}=0$, $\mu_{i,*}=-\dot{\lambda}_{i,*}$ or $\rho_{i,*} = -E_i^T \dot{x}_* - F \dot{ \lambda}_{i,*}$, $\mu_{i,*}=0$.

	Since the implications hold for almost all $t$, we conclude that $(x(t), \lambda(t), \dot{\lambda}(t)) \in \Gamma'_\text{SOL}$ for almost all $t \geq 0$. The result follows from \eqref{eq:ineq2}.
\end{IEEEproof}
Following Proposition \ref{constraints} and Theorem \ref{theorem_stability_nonunique}, if the matrix inequalities (\clr{see Appendix A for the full derivation}) over basic semialgebraic sets \eqref{eq:ineq1}, \eqref{eq:ineq2} are satisfied, one can conclude that the equilibrium $x_e$ is Lyapunov stable. Similarly, one can show that the equilibrium is exponentially stable if the left side of \eqref{eq:ineq2} is upper-bounded by $- \gamma_3 ||\boldsymbol{x}||_2^2$ as in Theorem \ref{theorem_stability_nonunique}.

\subsection{Control Design}
\rev{Now, the sets $\Gamma_{\text{SOL}}(E,F,c)$, $\Gamma'_{\text{SOL}}(E,F,c)$ are defined and it is assumed that there is access to a matrix $W$. The BMI feasibility problem with strictly positive constants $\gamma_1, \gamma_2$ and non-negative $\gamma_3$ can be formulated as:}
\begin{alignat}{2}
\label{eq:feasability_LCS}
& \underset{}{\text{find}} && V(\boldsymbol{x},\boldsymbol{\lambda}), K, L \\
\notag& \text{s.t.}  \quad && \gamma_1 ||\boldsymbol{x}||_2^2 \leq V(\boldsymbol{x},\boldsymbol{\lambda}) \leq \gamma_2 ||\boldsymbol{x}||_2^2, \; (\boldsymbol{x},\boldsymbol{\lambda}) \in \Gamma_{\text{SOL}}(E,F,c),\\
\notag& && \frac{dV}{dt} \leq -\gamma_3 ||\boldsymbol{x}||_2^2, \; (\boldsymbol{x},\boldsymbol{\lambda},\boldsymbol{\dot{\lambda}}) \in \Gamma'_{\text{SOL}}(E,F,c),
\end{alignat}
with the function $V(\boldsymbol{x},\boldsymbol{\lambda})$ as in \eqref{eq:lyap_nonunique} and
\begin{align*}
\frac{d V}{d t} & = 2 \boldsymbol{x}^T P (A \boldsymbol{x} + D \boldsymbol{\lambda} + a) + 2 (A \boldsymbol{x} + D \boldsymbol{\lambda} + a)^T \tilde{Q} W \boldsymbol{\lambda}\\
& + 2 \boldsymbol{x}^T \tilde{Q} W \dot{\lambda}  + 2 \boldsymbol{\lambda}^T W^T \tilde{R} W \boldsymbol{\dot{\lambda}} + p^T \dot{x} + \tilde{r} W_i^T \boldsymbol{\dot{\lambda}},
\end{align*}
\rev{with the controller as in \eqref{eq:cont_used}}.
Here, $V$ encodes the non-smoothness of the problem structure, mirroring the structure of the LCS, and allow tactile feedback design without exponential enumeration. This is an appealing middle ground between the common Lyapunov function of our prior work \cite{posa2015stability}, and purely hybrid approaches \cite{Papachristodoulou09a, marcucciwarm}. \clr{ Whereas methods like our prior work are more conservative than the proposed method (Example 3.3., \cite{camlibel2006lyapunov}), purely hybrid methods are less conservative at the cost of additional computation.}
\rev{ Here, it is possible to assign a different Lyapunov function and a control policy for each mode while avoiding mode enumeration so the approach can scale to large number of contacts $m$ unlike purely hybrid approaches.}

\clr{ Notice that the inequality with $\frac{dV}{dt}$ in \eqref{eq:feasability_LCS} is a bilinear matrix inequality because of the bilinear terms such as $PA$ where $A$ as in \eqref{eq:LCS} depends on the gain matrix $K$ as discussed in Section \ref{section:LCS_tactile}.} 
\rev { In \eqref{eq:feasability_LCS}, the problem of designing a control policy is formulated as finding a feasible solution for a set of bilinear matrix inequalities.}
The sets $\Gamma_{\text{SOL}}(E,F,c)$ and $\Gamma'_{\text{SOL}}(E,F,c)$ are incorporated via the S-procedure.

\subsection{Computing $W$ via Polynomial Optimization}
\rev{Until this point, it is assumed that there is access to a $W$ such that $W \text{SOL}(Ex+c,F)$ is a singleton for all $x \in \mathbb{R}^n$.  If $F$ is a P-matrix, one can always pick $W = I$ since $\text{SOL}(Ex+c,F)$ is a singleton for any $x$ as discussed earlier. For the non-P case, one can always trivially pick $W = 0$ which turns the Lyapunov function \eqref{eq:lyap_nonunique} into a common Lyapunov function, and controller \eqref{eq: controller} into a non-switching state feedback controller.} On the other hand, it is clearly better to search over a wider range of Lyapunov functions and controllers \cite{johansson1997computation}, \cite{camlibel2006lyapunov}. \rev{ Hence it is desired to maximize the rank of $W$. More precisely, consider the following optimization problem: }
\begin{alignat*}{2}
& \underset{W}{\text{max}} && \text{rank}(W) \\
& \text{subject to}  \quad && W \text{SOL}(q,F) \; \text{is a singleton for all} \; q. 
\end{alignat*}
\rev{To solve this problem, an algorithm based in a sequence of convex optimization problems is proposed. First, consider the following sub-problem:}
\clr{
	\begin{alignat}{2}
	\label{eq:subproblem_W}
	& \underset{w}{\text{find}} \quad \text{s.t.} \quad w^T \text{SOL}(q,F) \; \text{is a singleton for all} \; q,
	\end{alignat} 
	where $w \in \mathbb{R}^m$. Using this sub-problem, we will construct an algorithm to find matrices $W$. Notice that a $w$ such that $ |w^T (\lambda_{1,q} - \lambda_{2,q}) | = 0$ holds for all $q$, and all $\lambda_{1,q}, \lambda_{2,q} \in \text{SOL}(q,F)$ satisfies \eqref{eq:subproblem_W}.	Next, we demonstrate that it is sufficient to satisfy $|w^T (\lambda_{1,q} - \lambda_{2,q}) | \leq \eta$ for any $\eta > 0$ to satisfy  \eqref{eq:subproblem_W}.
	}
\begin{proposition}
	\label{unique_map}
	\clr{Suppose that for some $w$, the following inequalities hold for all $q$, all $\lambda_{1,q}, \lambda_{2,q} \in \text{SOL}(q,F)$:}
	\begin{equation}
	\begin{aligned}
	\label{eq:calculate_W}
	& ( \eta + w^T (\lambda_{1,q} - \lambda_{2,q}) )(\lambda_{1,q}^T \lambda_{1,q} + \lambda_{2,q}^T \lambda_{2,q}) \geq 0,\\
	&(\eta -w^T (\lambda_{1,q} - \lambda_{2,q}) ) (\lambda_{1,q}^T \lambda_{1,q} + \lambda_{2,q}^T \lambda_{2,q}) \geq 0,
	\end{aligned}
	\end{equation}
	where $\eta > 0$ is a constant slack parameter. Then, $w^T \text{SOL}(q,F)$ is a singleton for all $q$.
\end{proposition}
\begin{IEEEproof}
	Observe that
	\begin{equation*}
	\lambda \in \text{SOL}(q,F) \implies \alpha \lambda \in \text{SOL}(\alpha q,F),
	\end{equation*}
	for all $\alpha \geq 0$.
	We will show that the positive homogeneity property leads to:
	\begin{equation}
	\label{eq:slack_param}
	|w^T (\lambda_{1,q} - \lambda_{2,q})| \leq \eta \; \forall q \implies |w^T (\lambda_{1,q} - \lambda_{2,q})| = 0 \; \forall q.
	\end{equation}
	Assume that there exists $\eta^* > 0$ such that $|w^T (\lambda_{1,q^*} - \lambda_{2,q^*})| = \eta^*$ for some $q^*$. 
	Pick $\alpha^* > 0$ such that $\alpha^* \eta^* > \eta$ and $|w^T (\alpha^* \lambda_{1,q^*} - \alpha^* \lambda_{2,q*})| = \alpha^* \eta^* > \eta$.
	Due to the positive homogeneity property, there exists $\lambda_{1,\alpha q^*}$ and $\lambda_{2,\alpha q^*}$ such that $|w^T ( \lambda_{1,\alpha q^*} - \lambda_{2,\alpha q*})| = \alpha^* \eta^* > \eta$.
	This leads to a contradiction.	
	
	Next, we consider $q$ such that $(\lambda_{1,q}^T \lambda_{1,q} + \lambda_{2,q}^T \lambda_{2,q}) > 0$. 
	It follows from \eqref{eq:calculate_W} that $|w^T (\lambda_{1,q} - \lambda_{2,q})| \leq \eta$ hence $|w^T (\lambda_{1,q} - \lambda_{2,q})| = 0$. 
	If $(\lambda_{1,q}^T \lambda_{1,q} + \lambda_{2,q}^T \lambda_{2,q}) = 0$, then $\lambda_{1,q} = \lambda_{2,q} = 0$ and it trivially holds that $w^T \lambda_{1,q} = w^T \lambda_{2,q}$. 
	Therefore, for any $w$ such that \eqref{eq:calculate_W} holds, $w^T \lambda_{1,q} = w^T \lambda_{2,q}$ also holds for all $q$. Hence, $w^T SOL(q,F)$ is a singleton for all $q$.
\end{IEEEproof}
\clr{ \rev{ Finding a $w$ such that \eqref{eq:calculate_W} holds can be reduced to a polynomial optimization program (see Appendix B, \eqref{eq:find_W_poly_exp}) and any vector $w$ that satisfies \eqref{eq:calculate_W} also satisfies \eqref{eq:subproblem_W}. } 
Experimentally, we found that use of the slack variable $\eta$ was helpful to avoid numerical difficulties in the solvers.	
The solvers (Mosek \cite{mosek2010mosek}, SeDuMi \cite{sturm1999using}) had trouble verifying the status of the problem (feasible or infeasible) when $\eta = 0$.
The $(\lambda_{1,q}^T \lambda_{1,q} + \lambda_{2,q}^T \lambda_{2,q})$ terms in \eqref{eq:calculate_W} are introduced because there are degree two S-procedure terms (as shown in \eqref{eq:find_W_poly_exp}) and the inequalities must be at least degree two to use such S-procedure terms.
}
\begin{algorithm}[t]
	\caption{Find $W$}
	\begin{algorithmic}[1]
		\REQUIRE $F$	
		\\ \textit{Initialization} : $N \leftarrow I$, $W = [\;]$, $r = \mathbf{1}$, $w = \mathbf{1}$
		\WHILE {$\min r^T N^T w \neq 0$}
		\STATE $r \sim U(0,1)$
		\STATE Solve \eqref{eq:find_W_poly} and obtain $w$
		\IF {$\min r^T N^T w < 0$}
		\STATE $W \leftarrow \begin{bmatrix}
		W \\ w^T
		\end{bmatrix}$
		\STATE Calculate $N$ based on $\mathcal{N}(W)$
		\ENDIF
		\ENDWHILE
		\RETURN $W$ 
	\end{algorithmic} 
	\label{algortihm_findW}
\end{algorithm}
Given a matrix $W_d \in \mathbb{R}^{s \times m}$, one can utilize Proposition \ref{unique_map} in order to find a vector $w$ such that $w^T \text{SOL}(q,F)$ is a singleton (for all $q$) and $w^T$ is linearly independent with the rows of $W_d$. Consider the optimization problem:
\begin{alignat}{2}
\label{eq:find_W_poly}
& \underset{w, \eta}{\text{min}} && r^T N^T w \\
\notag& \text{subject to}  \quad && (\eta + w^T (\boldsymbol{\lambda}_1 - \boldsymbol{\lambda}_2)) (\boldsymbol{\lambda}_1^T \boldsymbol{\lambda}_1 + \boldsymbol{\lambda}_2^T \boldsymbol{\lambda}_2) \geq 0, \\
\notag& &&  (\eta -w^T (\boldsymbol{\lambda}_1 - \boldsymbol{\lambda}_2)) (\boldsymbol{\lambda}_1^T \boldsymbol{\lambda}_1 + \boldsymbol{\lambda}_2^T \boldsymbol{\lambda}_2) \geq 0,  \\
\notag& && \text{for} \; \boldsymbol{\lambda}_1, \boldsymbol{\lambda}_2 \in \text{SOL}(\boldsymbol{q},F), \\
\notag& && |w_i| \leq 1, \; \forall i, \; \eta \geq 0,
\end{alignat}
where $r$ is a random vector with entries sampled from uniform distribution ($r_i \sim U(0,1))$, $N$ is a basis for the nullspace of $W_d$ ($\mathcal{N}(W_d)$), $\boldsymbol{\lambda}_1, \boldsymbol{\lambda}_2, \boldsymbol{q}$ are indeterminates, and the set inclusion is incorporated via the S-procedure for the first two inequalities \clr{(Appendix B, \eqref{eq:find_W_poly_explicit})}. \rev{ Randomness is introduced in \eqref{eq:find_W_poly} in order to ensure that the objective function is almost surely strictly negative. More precisely, a linear objective that is strictly negative if any $N$ such that $N w \neq 0$ exists is needed and projection onto a random vector $r$ ensures this with probability one.}
\begin{proposition}
	\label{find_W_proposition}
	Consider $W_d$ and the optimization \eqref{eq:find_W_poly}. If there exists a $w$ such that the constraints hold, and $w^T$ is linearly independent with the rows of $W_d$, then $\min r^T N^T w < ~ 0$ almost surely.
\end{proposition}
\begin{IEEEproof}
	Assume there exists a $w$ that is feasible for optimization problem \eqref{eq:find_W_poly} and $w^T$ is linearly independent with the rows of $W_d$. Then, $||N^T w|| > 0$ and $r^T N^T w \neq 0$ with probability 1. By homogeneity, an optimal $w_*$ can be found such that $r^T N^T w_* < 0$.
\end{IEEEproof}
We now introduce Algorithm \ref{algortihm_findW} based on Proposition \ref{find_W_proposition}.
The algorithm almost surely finds a new linearly independent vector that satisfies the constraints in \eqref{eq:find_W_poly} if it exists and terminates when there are not any left. \clr{Notice that the randomization process affects the outcome ($W$) but we are only interested in finding a $W$ such that $W \text{SOL}(q,F)$ is a singleton for all $q$ hence this effect can be neglected. }

\begin{figure}[t]
	\label{fig:Cart-pole with soft walls}
	\includegraphics[width=1.1\columnwidth]{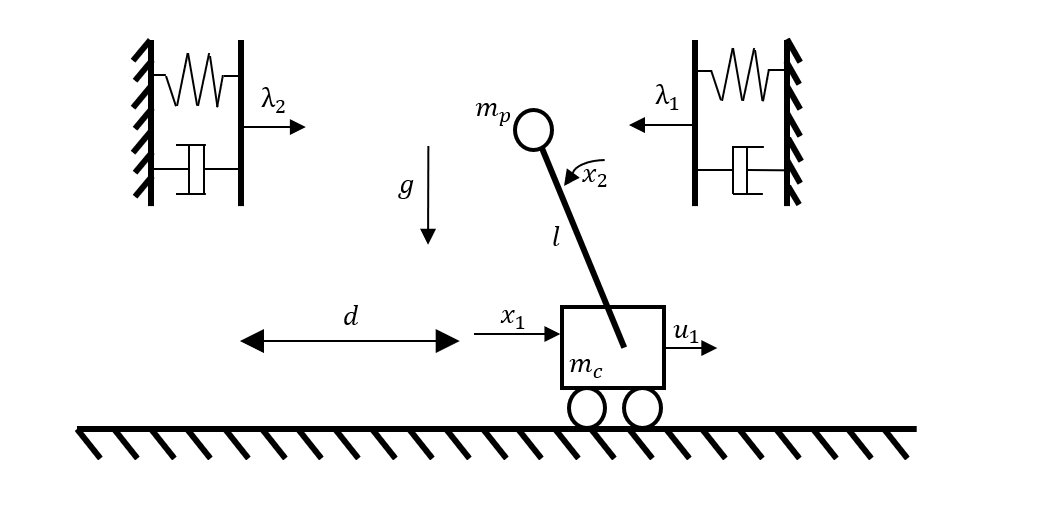}
	\caption{Benchmark problem: Regulation of the cart-pole system to the origin with soft walls.}
	\label{fig:cartpole_fig}
\end{figure}

\section{Numerical Examples}
\label{section:examples}
\rev{ In this paper, the YALMIP \cite{Lofberg2004} toolbox and PENBMI \cite{kovcvara2003pennon} are used to formulate and solve bilinear matrix inequalities. SeDuMi \cite{sturm1999using} and Mosek \cite{mosek2010mosek} are used for solving the semidefinite programs (SDPs).} PATH \cite{dirkse1995path} has been used to solve the linear complementarity problems when performing simulations. \rev{MATLAB's stiff solver 'ode15s' is used while performing simulations of the LCS models and Euler's method with stepsize $10^{-4}$ is used for the nonlinear models.} The code for all examples is available\footnote{ \url{https://github.com/AlpAydinoglu/cdesign}} and examples are provided with a video depiction\footnote{ \url{https://www.youtube.com/watch?v=CpYZcinYuQM} }. The experiments are done on a desktop computer with the processor \textit{Intel i7-9700} and \textit{16GB RAM}. \rev{ We have reported the offline computation times for all of the examples and emphasize that our controller only requires only a few addition and multiplication operations when running online and is applicable in real time context after the offline computations are done. }

\subsection{Cart-Pole with Soft Walls}
\label{subsection:cartpole}
\rev{ Consider the cart-pole system where the goal is to balance the pole and regulate the cart to the center, where there are frictionless walls, modeled via spring contacts, on both sides. }
This problem, or a slight variation of it, has been used as a benchmark in control through contact \cite{marcucciwarm}, \cite{deits2019lvis}, \cite{marcucci2017approximate} and the model is shown in Figure \ref{fig:cartpole_fig}.

\begin{figure}[t]
	\hspace*{-0.9cm}
	\includegraphics[width=1.2\columnwidth]{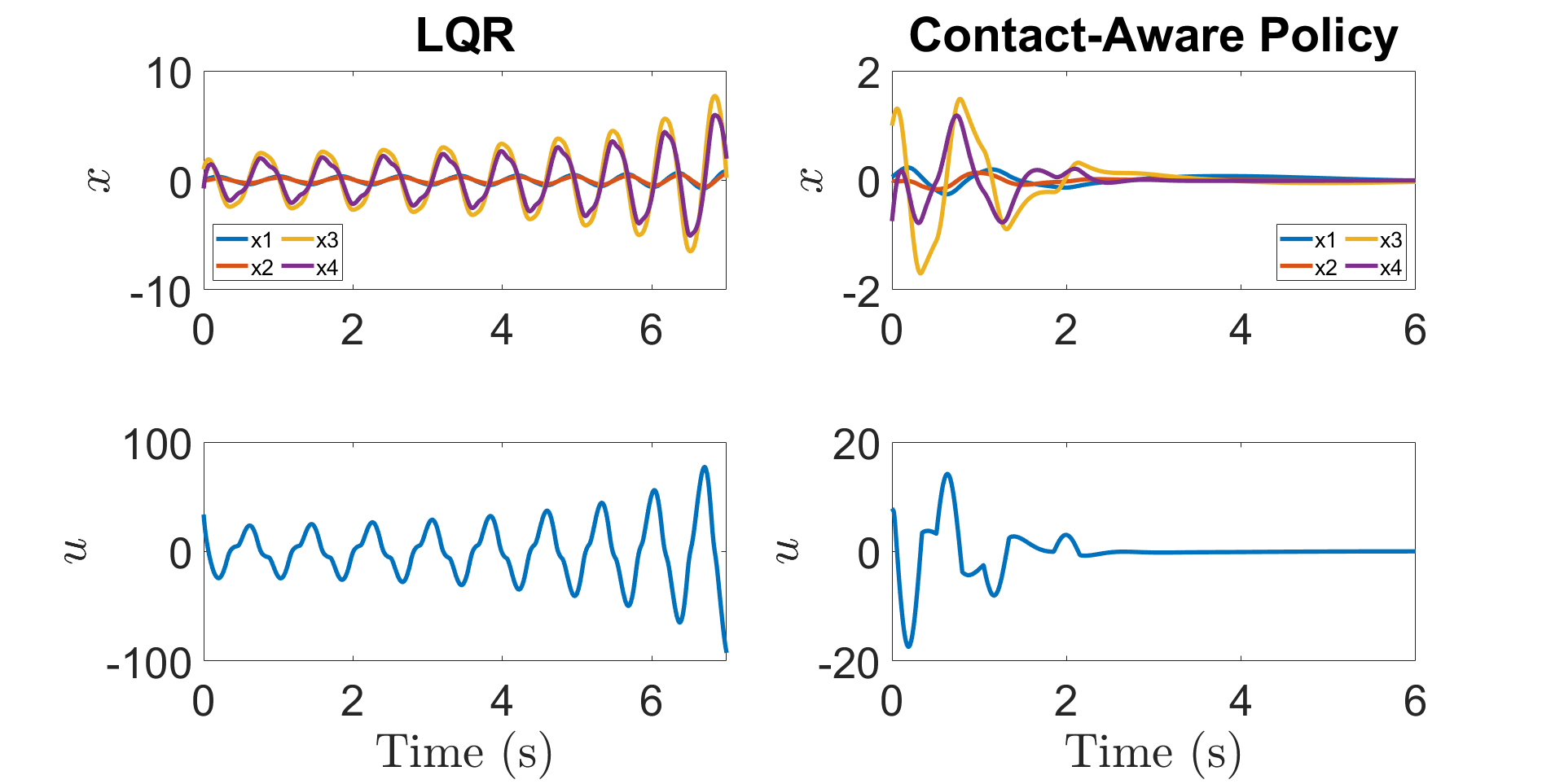}
	\caption{Performance of LQR and contact-aware policy starting from the same initial condition for the cart-pole with soft walls example. LQR is unstable whereas contact-aware policy is successful.}
	\label{carta}
\end{figure}

\begin{figure}[b]
	\hspace*{-1cm}
	\includegraphics[width=1.2\columnwidth]{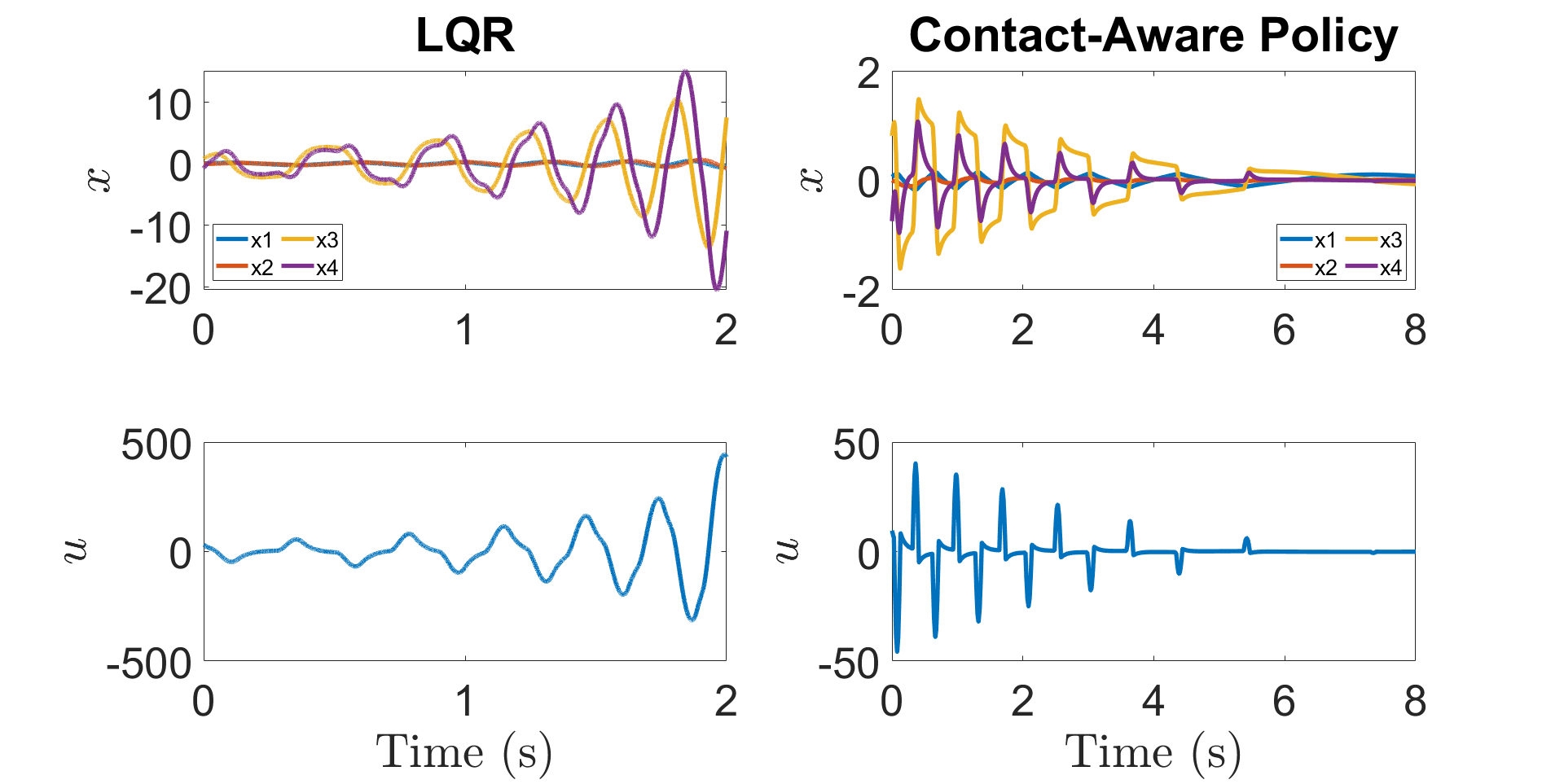}
	\caption{Comparison of the LQR controller and contact-aware policy for $k_1 ~= ~k_2 = 100$. The LQR controller fails to stabilize the system whereas contact-aware policy is successful.}
	\label{carta_100}
\end{figure}

\rev{ First, the model where there is no damping is analyzed.} In this model, the $x_1$ is the position of the cart, $x_2$ is the angle of the pole, and $x_3$, $x_4$ are their respective time derivatives. 
The input $u_1$ is a force applied to the cart, and the contact forces of the walls are represented with $\lambda_1$ and $\lambda_2$, leading to the LCS
\begin{align}
    \label{eq:cartpole_dyn_1}
    \dot{x}_1& = x_3, \\
    \dot{x}_2& = x_4, \\
    \dot{x}_3 &= \frac{g m_p}{m_c} x_2 + \frac{1}{m_c} u_1,\\
    \label{eq:cartpole_dyn_2}
    \quad \dot{x}_4 &= \frac{g (m_c + m_p)}{l m_c}  x_2 + \frac{1}{l m_c} u_1 + \frac{1}{l m_p} \lambda_1 - \frac{1}{l m_p} \lambda_2,\\
    \notag \quad 0 &\leq \lambda_1 \perp l x_2 - x_1 + \dfrac{1}{k_1}  \lambda_1 + d \geq 0,\\
    \notag 0 &\leq \lambda_2 \perp x_1 - l x_2 + \dfrac{1}{k_2}  \lambda_2 + d \geq 0,
\end{align}
where $k_1 = k_2 = 10$ are stiffness parameters of the soft walls, $g = 9.81$ is the gravitational acceleration, $m_p = 0.1$ is the mass of the pole, $m_c = 1$ is the mass of the cart, $l = 0.5$ is the length of the pole, and $d = 0.1$ represents where the walls are. \rev{Observe that the model has absolutely continuous solutions following Proposition \ref{existence_uniqueness}.} For this model, we solve the feasibility problem \eqref{eq:feasability_LCS} and find a controller of the form $u(x,\lambda) = Kx + L \lambda$ that regulates the model to the origin \clr{with $K = \begin{bmatrix}
	3.69 & -46.7 & 3.39 & -5.71
	\end{bmatrix}$ and $L = \begin{bmatrix}
	-13.98 & 13.98
	\end{bmatrix}$.}
The algorithm succeeded in finding a feasible controller in 0.72 seconds. \clr{Additionally, we have tried to find a pure state feedback controller ($L=0$) and, as formulated, failed to find such a controller for 1000 trials starting from different initial conditions. 
\rev { Due to the non-convexity of the BMI, this does not guarantee that such a controller-Lyapunov function pair does not exist, but demonstrates that the optimization problem is harder to solve in more conservative settings. }
}
\begin{figure}[t]
	\hspace*{-0.4cm}
	\includegraphics[width=1.2\columnwidth]{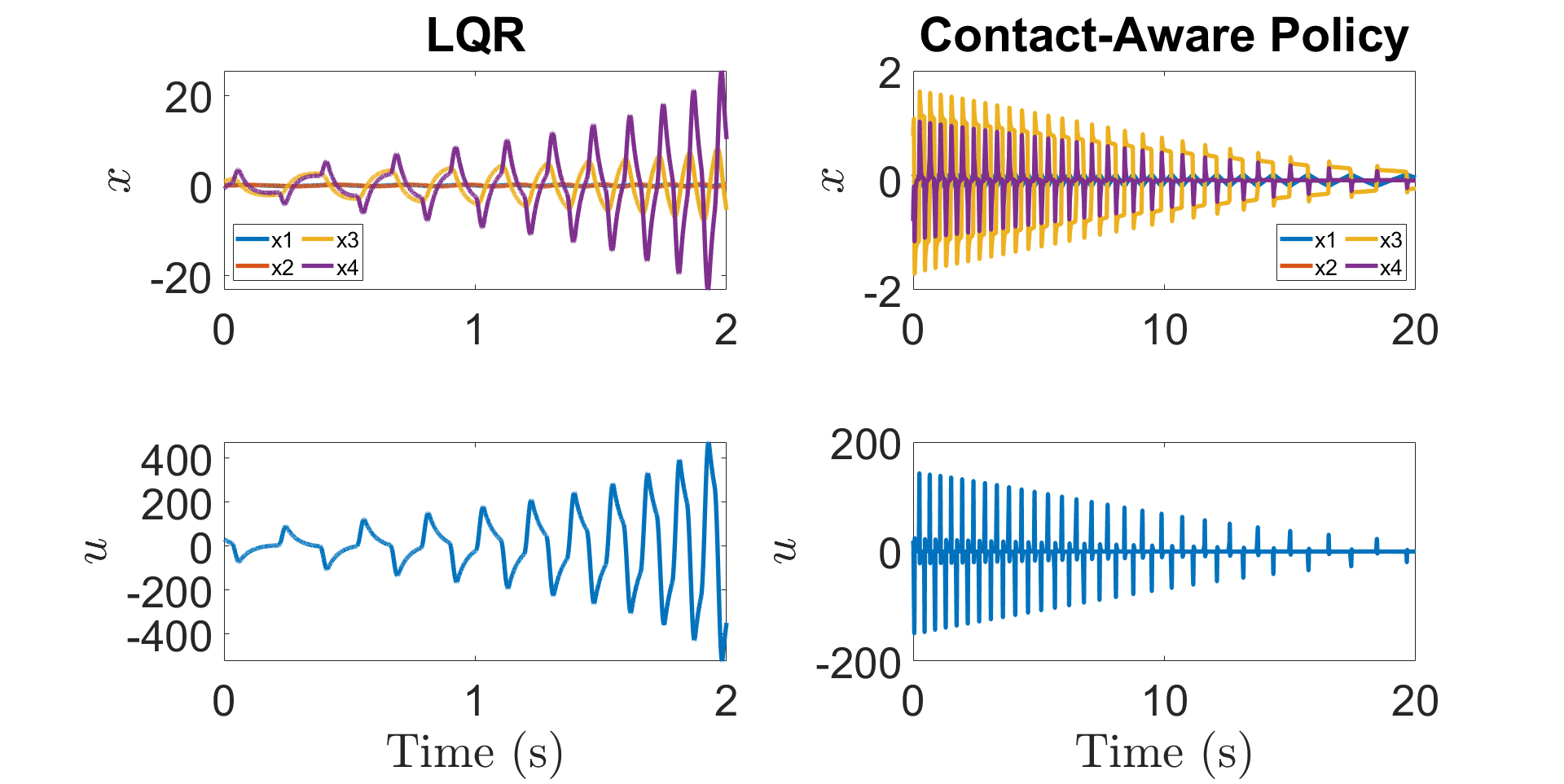}
	\caption{Comparison of the LQR controller and contact-aware policy for $k_1 ~= ~k_2 = 1000$. The LQR controller fails to stabilize the system whereas contact-aware policy is successful.}
	\label{carta_1000}
\end{figure}

\begin{figure}[b!]
	\hspace*{-0.4cm} 
	\includegraphics[width=1.1\columnwidth]{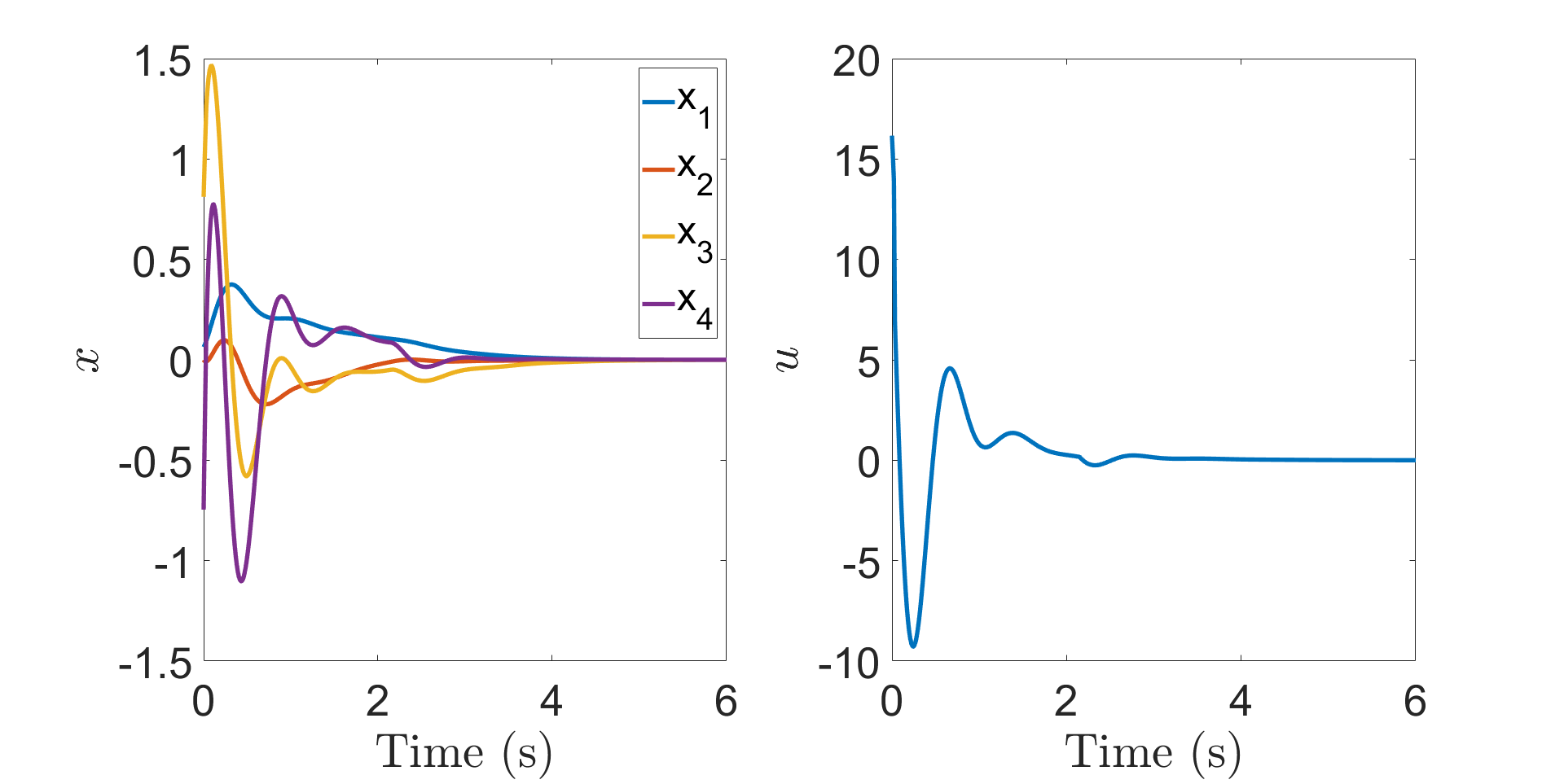}
	\caption{Performance of tactile feedback controller with damping. Contact-aware policy successfully stabilizes the nonlinear plant.}
	\label{fig:damping}
\end{figure}

\begin{figure*}[t]
	\includegraphics[width=2\columnwidth]{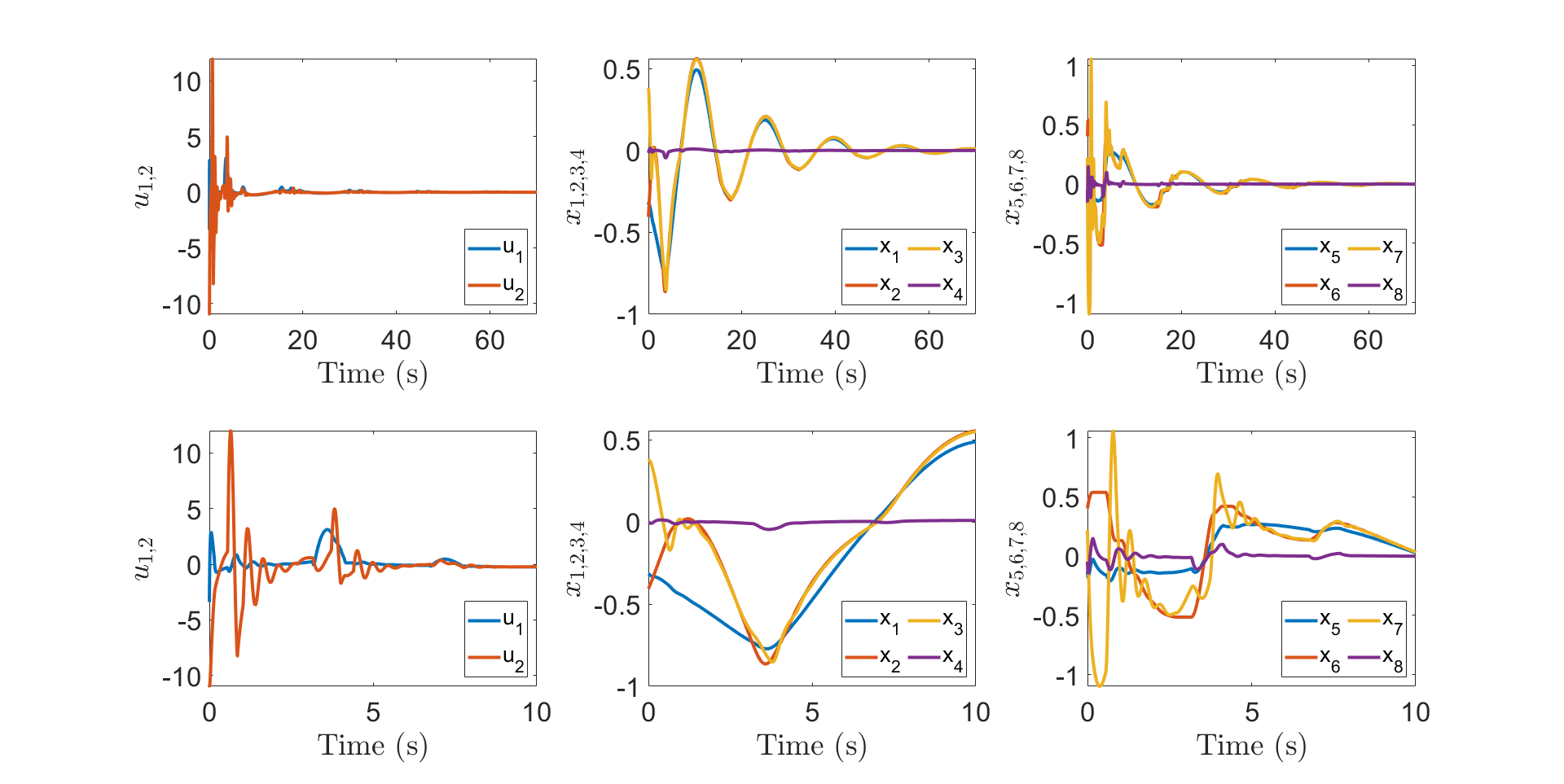}
	\caption{Simulation with contact-aware policy for partial state-feedback example. The plots on the top row show the input and the state variables ($u(t),x(t)$) for the time interval $t=[0 \; 60]$. Second row demonstrates the time interval $t=[0 \; 10]$ for the same initial condition. }
	\label{fig:partial}
\end{figure*}

\rev{ As a comparison, consider an LQR controller with penalty on the state $Q = 100 I$ and penalty on the input $R=1$ \clr{which is given as $K_\text{LQR} ~=~ \begin{bmatrix} 10 & -91.77 & 16.28 & -22.69
	\end{bmatrix} $.}} \rev{ Both contact-aware and LQR controllers are tested on the nonlinear plant for 100 initial conditions where $x_2(0)=0$, and $x_1(0), x_3(0), x_4(0)$ are uniformly distributed ($10x_1(0), \frac{1}{4}x_3(0), x_4(0) \sim U[-1, 1]$).} The LQR controller was successful only 31\% of the time, whereas our contact-aware policy was successful 87\% of the time. \rev { In Figure \ref{carta}, an example is presented where both LQR and contact-aware policy start from the same initial conditions and LQR fails whereas our policy is successful.} \rev{
	Our method is compared with LQR, which uses linearization and thus ignores potential contact events. Our controller is a contact-aware analogue to non-switching state-feedback controllers such as LQR. In our comparison, the LQR controller acts as a stand-in for these methods which do not explicitly consider the non-smooth structure of the system.}

\rev{Then, two different cases with higher stiffness values are explored. First, a controller is designed for the case where $k_1 = k_2 = 100$ with the controller gains:
\begin{align*}
	& K_{100} = \begin{bmatrix}
	3.69 & -48.78 & 2.36 & -9.96
	\end{bmatrix}, \\
	& L_{100} = \begin{bmatrix}
	-14.14 & 14.14 \end{bmatrix}.
\end{align*}
The performance of the contact-aware controller is demonstrated against the previously designed LQR on the nonlinear plant in Figure \ref{carta_100}.
}

\rev{One more set of experiments is presented where $k_1 = k_2 = 1000$ and a controller is designed with the gains: 
\begin{align*}
& K_{1000} = \begin{bmatrix}
0.45 & -40.23 & 0.86 & -25.50
\end{bmatrix}, \\
& L_{1000} = \begin{bmatrix}
-14.14 & 14.14 \end{bmatrix}.
\end{align*}
The performance of the contact-aware controller is demonstrated against the previously designed LQR on the nonlinear plant in Figure \ref{carta_1000}. Notice that it is possible to design controllers for stiffer contacts, but
stiff, near-impulsive contact events resolve very quickly and that measurement and rapid response may not be practical.
}

Next, inspired by \cite{brogliato2003some}, consider the case where damping term, $b$, is nonzero. The system dynamics are modeled as in \eqref{eq:cartpole_dyn_1}-\eqref{eq:cartpole_dyn_2} with the following complementarity constraints:
	\begin{align*}
	& 0 \leq \lambda_1 \perp - k x_1 + k l x_2 - b x_3 + bl x_4 + kd + \lambda_1 + \gamma_1 \geq 0, \\
	& 0 \leq \gamma_1 \perp M x_1 - M l x_2 - M d + \gamma_1 \geq 0, \\
	& 0 \leq \lambda_2 \perp k x_1 - l k x_2 + b x_3 - b l x_4 + kd + \lambda_2 + \gamma_2 \geq 0, \\
	& 0 \leq \gamma_2 \perp -M x_1 + M l x_2 - Md + \gamma_2,
	\end{align*}
	where a large positive constant $M=1000$ is introduced and the slack variables $\gamma_1$ and $\gamma_2$ to capture the affect of damper. \rev{As $M$ increases the model approximates the spring-damper model \cite{brogliato2003some} better.} In this model, consider $b=1$, $k=10$, $m_p = m_c = 1$, $g=9.81$, $l=0.5$, $d=0.1$. For this model, we solve the feasibility problem \eqref{eq:feasability_LCS} and find a controller of the form $u(x,\lambda) = Kx + L \lambda$ that regulates the model to the origin with $K = \begin{bmatrix}
	8.47 & -64.54 & 10.36 & -9.69
	\end{bmatrix}$ and $L = \begin{bmatrix}
	-4.8 & 0 & 4.76 & 0
	\end{bmatrix}$ in 384 seconds. The performance of the controller is demonstrated on the nonlinear plant in Figure \ref{fig:damping}.

\subsection{Partial State Feedback}
\begin{figure}[t!]
	\hspace*{0.8cm}\includegraphics[width=0.8\columnwidth]{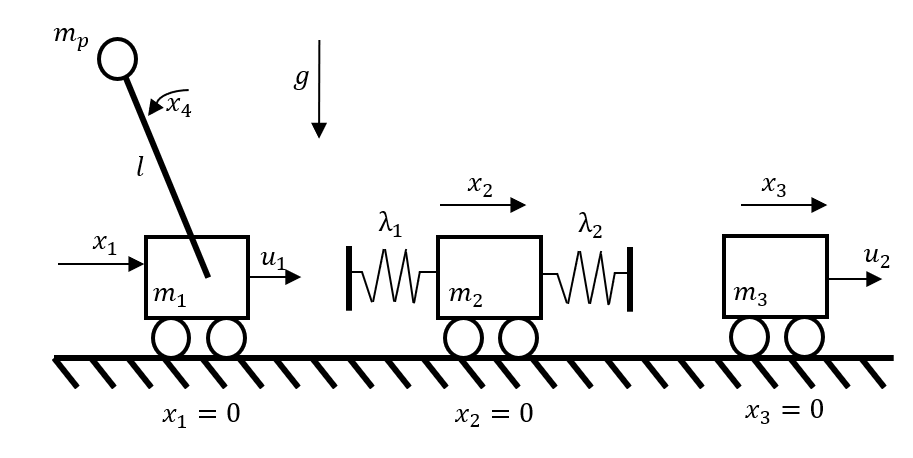}
	\caption{Regulation of carts to their respective origins without  observation of the middle cart.}
	\label{threecarts}
\end{figure}

\rev{ Consider a model that consists of three carts on a frictionless surface as in Figure \ref{threecarts}. } The cart on the left is attached to a pole and the cart in the middle makes contact via soft springs. In this model, a spring only becomes active if the distance between the outer block and the block in the middle is less than some threshold. Here, $x_1,x_2,x_3$ represent the positions of the carts and $x_4$ is the angle of the pole. 
The corresponding LCS is
\begin{align*}
    \ddot{x}_1 &= \frac{g m_p}{m_1} x_4 + \frac{1}{m_1} u_1 - \frac{1}{m_1} \lambda_1, \\
    \ddot{x}_2 &= \frac{\lambda_1}{m_2} - \frac{\lambda_2}{m_2}, \\
    \ddot{x}_3 &= \frac{\lambda_2}{m_3} + \frac{u_2}{m_3},\\
    \ddot{x}_4 &= \frac{g(m_1+m_p)}{m_1 l} x_4 + \frac{u_1}{m_1 l} - \frac{1}{m_1 l} \lambda_1,\\
    \notag 0 &\leq \lambda_1 \perp x_2 - x_1 + \frac{1}{k_1} \lambda_1 \geq 0,\\
    \notag 0 &\leq \lambda_2 \perp x_3 - x_2 + \frac{1}{k_2} \lambda_2 \geq 0,
\end{align*}
where the masses of the carts are $m_1 = m_2 = m_3 = 1$, $g = 9.81$ is the gravitational acceleration, $m_p = 1.5$ is the mass of the pole, $l = 0.5$ is the length of the pole, and $k_1 = k_2 = 20$ are stiffness parameters of the springs. 
Observe that we have control over the outer blocks, but do not have any control over the block in the middle. \rev{ Additionally, it is assumed that the middle block is not observed, and one can only observe the outer blocks and the contact forces. } \rev{Notice that the model has absolutely continuous solutions following Proposition \ref{existence_uniqueness}.}

\rev{ For this example, the feasibility problem \eqref{eq:feasability_LCS} takes 9.3 seconds to solve with a controller of the form } $u(x,\lambda) = Kx + L \lambda$ \clr{where
\begin{align*}
& K = \begin{bmatrix}
-2.8 & 0 & 6.6 & -263.1 & 6.4 & 0 & -2.1 & -30.2 \\ 11.5 & 0 & -12.1 & 12.1 & 2.6 & 0 & -4.7 & 6.6 
\end{bmatrix}, \\
& \qquad \qquad \qquad \qquad L = \begin{bmatrix}
-3.7 & -0.6 \\ -0.6 & 7.2
\end{bmatrix}.
\end{align*}
}
Notice that we enforce sparsity on the controller $K$ and do not use any feedback from the state $x_2$ or its derivative $\dot{x}_2$. \rev{ This example demonstrates that tactile feedback can be used in scenarios where full state information is lacking and also impact events can be used in order to stabilize the system. } \rev{ In Figure \ref{fig:partial}, the performance of the controller is demonstrated. }

\subsection{Acrobot with Soft Joint Limits}
\begin{figure}[t!]
	\hspace*{1cm} \includegraphics[width=0.8\columnwidth]{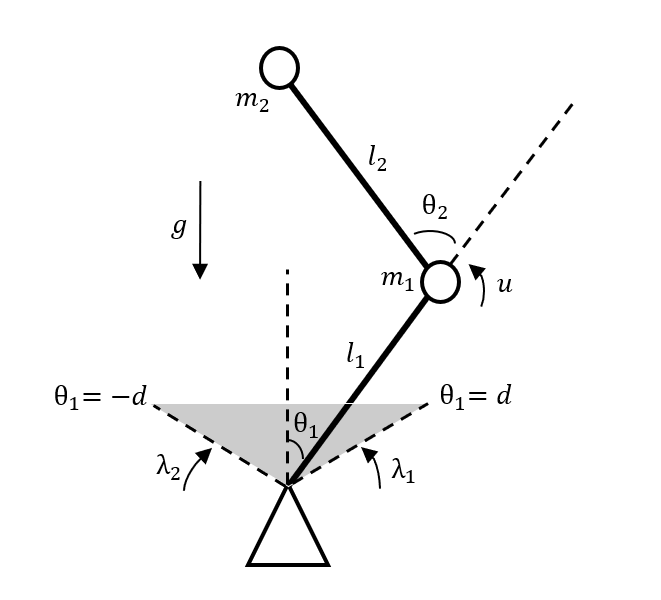}
	\caption{Acrobot with soft joint limits.}
	\label{fig:acrobot_fig}
\end{figure}

\rev{ As a third example, consider the classical underactuated acrobot, a double pendulum with a single actuator at the elbow (see \cite{murray1991case} for the details of the acrobot dynamics). Additionally, soft joint limits are added to the model. Hence we consider the model in Figure \ref{fig:acrobot_fig}: }
\begin{equation*}
	\dot{x} = A x + Bu + D \lambda,
\end{equation*}
where $x = (\theta_1, \theta_2, \dot{\theta}_1,\dot{\theta}_2)$, $\lambda = (\lambda_1, \lambda_2)$, and $D = \begin{bmatrix} 0_{2 \times 2} \\ M^{-1} J^T \end{bmatrix}$ with $J^T = \begin{bmatrix} -1 & 1 \\ 0 & 0 \end{bmatrix}$. For this model, the masses of the rods are $m_1 = 0.5$, $m_2 = 1$, the lengths of the rods are $l_1 = 0.5$, $l_2 = 1$, and the gravitational acceleration is $g = 9.81$. \rev{ The soft joint limits are modeled using the following complementarity constraints:}
\begin{align*}
	&0 \leq d - \theta_1 + \dfrac{1}{k} \lambda_1 \perp \lambda_1 \geq 0,\\
	&0 \leq \theta_1 + d + \dfrac{1}{k} \lambda_2 \perp \lambda_2 \geq 0,
\end{align*}
where $k = 1$ is the stiffness parameter and $d = 0.2$ is the angle that represents the joint limits in terms of the angle $\theta_1$. \clr{Observe that the model has absolutely continuous solutions, $x(t)$, following Proposition \ref{existence_uniqueness} since $F$ is a P-matrix.}

\begin{figure}[b!]
	\hspace*{-0.4cm}
	\includegraphics[width=1.2\columnwidth]{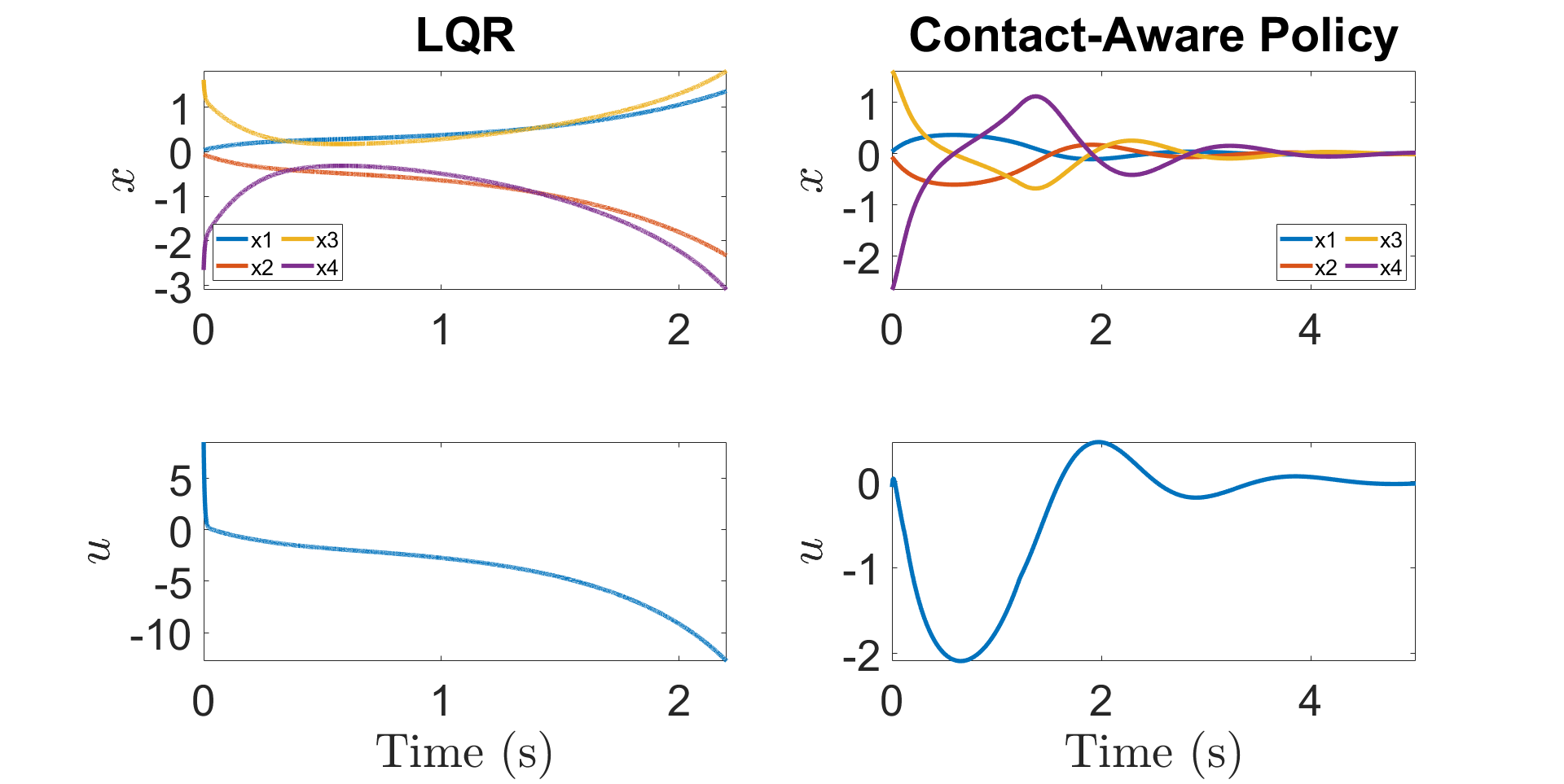}
	\caption{Simulation of LQR and contact-aware policy starting from the same initial condition for the acrobot with soft joint limits example. LQR is unstable whereas contact-aware policy is successful.}
	\label{acro_sim}
\end{figure}

For this example, we solve the feasibility problem \eqref{eq:feasability_LCS} and obtain a controller of the form $u(x,\lambda) = Kx + L \lambda$ in 1.18 seconds \clr{with $K = \begin{bmatrix}
	73.07 & 38.11 & 30.41 & 18.95
	\end{bmatrix}$ and $L = \begin{bmatrix}
	-4.13 & 4.13
	\end{bmatrix}$.}
For comparison, we also designed an LQR controller for the linear system where the penalty on the state is $Q=100I$ and the penalty on the input is $R=1$ \clr{which is given as $K_\text{LQR} ~=~ \begin{bmatrix} 1476.3 & 851.68 & 548.81 & 334.43
\end{bmatrix} $.} \rev{ 100 trials were made on the nonlinear plant where initial conditions were sampled according to $x_1(0) = x_2(0) = 0$ and ${x_3(0),x_4(0) \sim U[-0.05, 0.05]}$.} Out of these 100 trials, LQR was successful only 49\% of the time whereas our design was successful 87\% of the time. \rev{ In Figure \ref{acro_sim}, a case where LQR fails and contact-aware policy is successful is presented. }

\subsection{Box with Friction}
\label{sub:friction}
\begin{figure}[t!]
	\hspace*{2cm}
	\includegraphics[width=0.5\columnwidth]{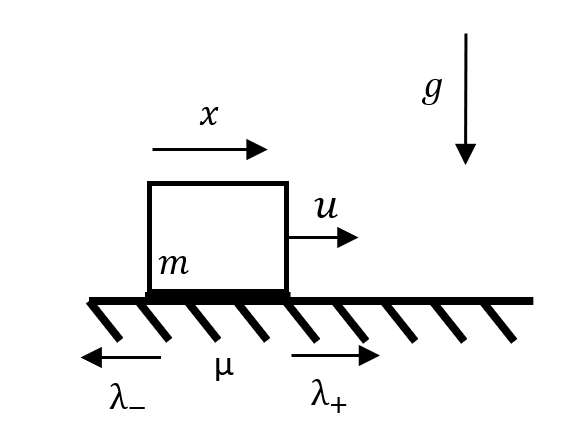}
	\caption{Regulation task of a box standing on a surface with Coulomb friction.}
	\label{fig:box_with_friction}
\end{figure}

\rev{ Consider a quasi-static model of a box on a surface, as in Figure \ref{fig:box_with_friction}, where $\mu$ is the coefficient of friction between the box and the ground. }
\rev{ Newtons's second law is approximated with a force balance equation with Coulomb friction and damping. }
The goal is to regulate the box to the center.
This simple model serves as an example where $F$ is not a P-matrix and the complementarity constraints have a dependency on the input $u$ ($H \neq 0$).
Here, $x$ is the position of the box, $u$ is the input, $\lambda_+$ is the positive component of the friction force, $\lambda_{-}$ is the negative component of the friction force and $\gamma$ is the slack variable:
\begin{align*}
& \alpha \dot{x} = u + \lambda_+ - \lambda_-, \\
&  0 \leq \gamma \perp \mu m g - \lambda_+ - \lambda_- \geq 0, \\
& 0 \leq \lambda_+ \perp \gamma + u + \lambda_+ -\lambda_- \geq 0, \\
& 0 \leq \lambda_- \perp \gamma - u -\lambda_+ + \lambda_-  \geq 0,
\end{align*}
where $m=1$ is the mass of the box, $g=9.81$ is the gravitational acceleration $\mu=0.1$ is the friction coefficient, and $\alpha=4$ is the damping coefficient. \rev{ Input delay is modeled with the low-pass filter:}
\begin{figure}[b]
	\hspace*{-0.5cm}
	\includegraphics[width=1.1\columnwidth]{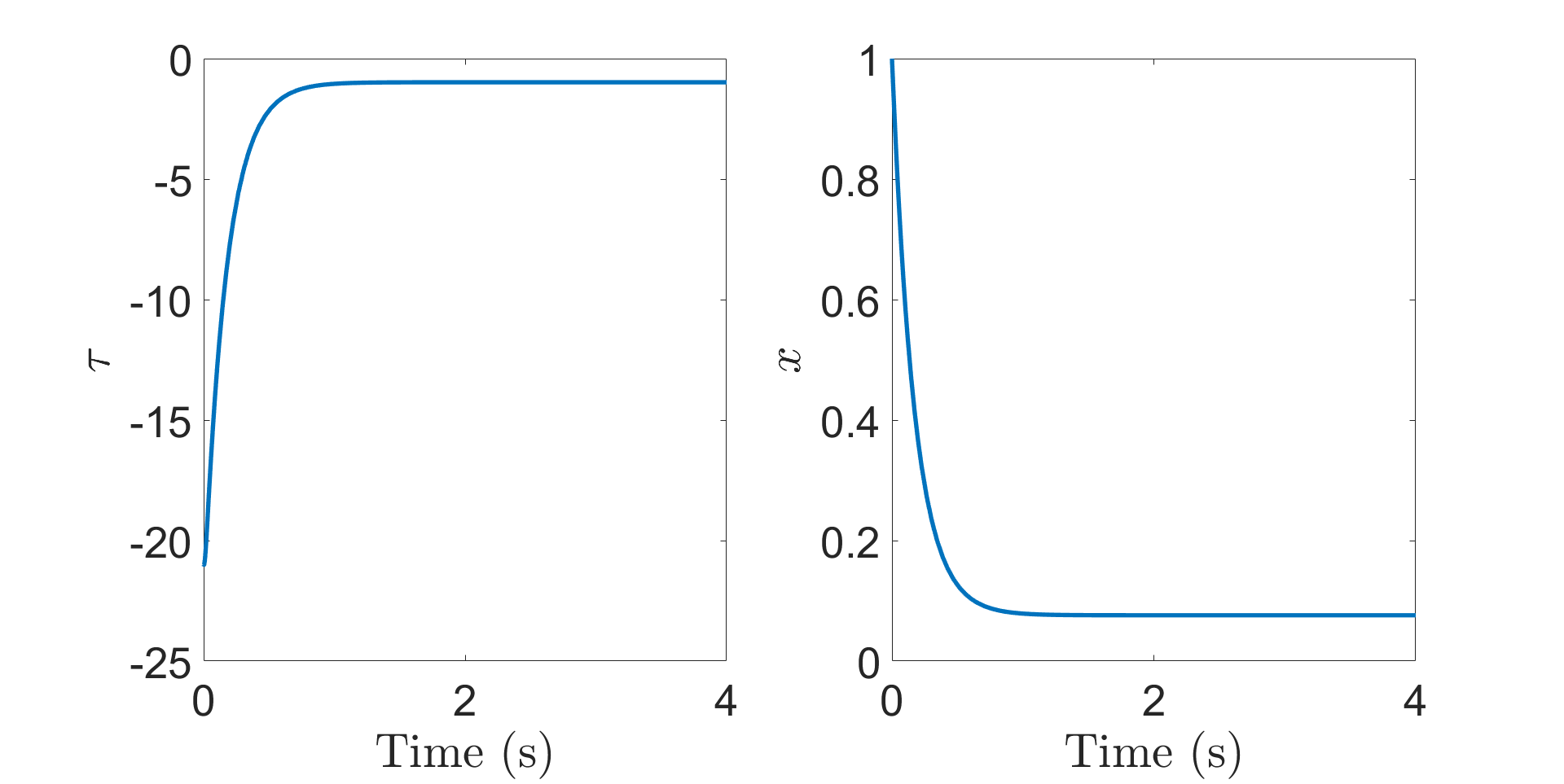}
	\caption{Simulation of box with friction example. The equilibrium is Lyapunov stable and the state trajectory, $x(t)$ does not reach origin because of stiction.}
	\label{fig:box_friction_sim}
\end{figure}
\begin{align*}
& \alpha \dot{x} = \tau + \lambda_+ - \lambda_-, \\
& \dot{\tau} = \kappa (u - \tau), \\
&  0 \leq \gamma \perp \mu m g - \lambda_+ - \lambda_-  \geq 0, \\
& 0 \leq \lambda_t^+ \perp \gamma + \tau + \lambda_+ -\lambda_- \geq 0, \\
& 0 \leq \lambda_t^- \perp \gamma - \tau -\lambda_+ + \lambda_-  \geq 0,
\end{align*}
where $\kappa = 100$. Since $F$ is not a P-matrix, we use Algorithm \ref{algortihm_findW} and find $W = [0 \; 1 \; -1]$ such that $W \text{SOL}(q,F)$ is a singleton for all $q$ \clr{in 8.09} seconds. For this example, $W$ shows that the net friction force, $\lambda_+ - \lambda_-$ is always unique. Notice that the x-trajectory, $x(t)$ is unique, but the $\lambda$-trajectory, $\lambda(t)$ is not.
\rev{Note that the closed-loop system has absolutely continuous solutions following Proposition \ref{existence_uniqueness} since it is enforced that $L \text{SOL}(Ex+c,F)$ is a singleton.}

We can solve the feasibility problem in \eqref{eq:feasability_LCS} in 22 seconds and find a controller of the form $u(x,\lambda) = Kx + L \lambda$ such that the system is Lyapunov stable \clr{with $K = \begin{bmatrix}
	-10.58
	\end{bmatrix}$ and $L = \begin{bmatrix}
	0 & 0.7 & -0.7
	\end{bmatrix}$.} \rev{ In Figure \ref{fig:box_friction_sim}, the performance of the controller is demonstrated.}

\subsection{Three Legged Table}
\label{sub:friction2}
We examine a variation of Example D and consider a three legged table on a surface with Coulomb friction as in Figure \ref{fig:3legged_table}.
In this model, the coefficient of friction values $(\mu_1, \mu_2, \mu_3)$ are different for each leg of the table. 
The normal forces at the legs of the table are denoted by $(N_1, N_2, N_3)$ and sum of the normal forces are equal to the mass times gravitational acceleration, $mg$.
The net friction force is unique in static situations but it is non-unique during sliding since individual normal forces, $N_i$, are non-unique.
The task is regulating the three legged table to the center. Newton's second law is approximated with a force balance equation with Coulomb friction and damping as in the previous example. In this model $x$ is the position of the box, $\tau$ is the output of the low-pass filter, $\lambda_+$ is the positive component of the friction force, $\lambda_-$ is the negative component of the friction force, $\gamma$ is the slack variable, $u$ is the force applied to the table:
\begin{figure}[t!]
	\hspace*{1cm}
	\includegraphics[width=0.7\columnwidth]{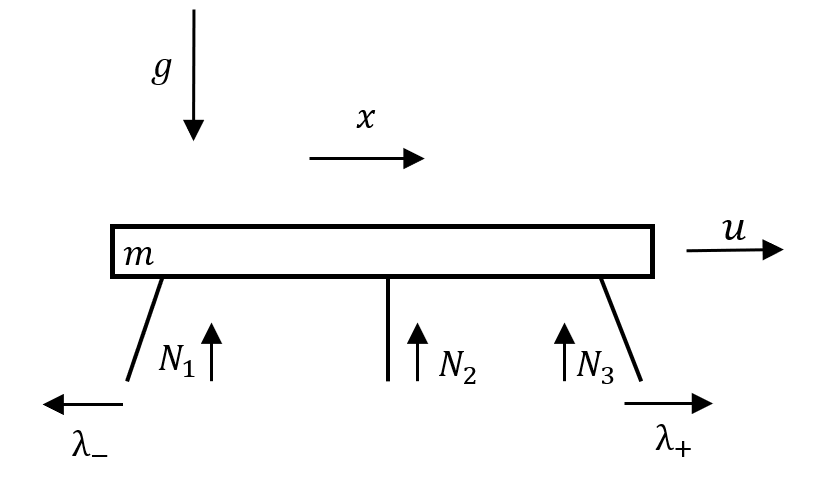}
	\caption{Regulation task of a 3 legged table.}
	\label{fig:3legged_table}
\end{figure}
\begin{align}
\label{eq:model11}
& \alpha \dot{x} = \tau + \lambda_+ - \lambda_-, \\
& \dot{\tau} = \kappa (u - \tau), \\
&  0 \leq \gamma \perp \mu_1 N_1 + \mu_2 N_2 + \mu_3 N_3 - \lambda_+ - \lambda_- \geq 0, \\
& 0 \leq \lambda_+ \perp \gamma + \tau + \lambda_+ -\lambda_- \geq 0, \\
& 0 \leq \lambda_- \perp \gamma - \tau -\lambda_+ + \lambda_-  \geq 0, \\
\label{eq:model33}
& N_1 + N_2 + N_3 = mg, \\
\label{eq:model22}
& N_1, N_2, N_3 \geq 0,
\end{align}
where $(\mu_1, \mu_2, \mu_3) = (0.1, 0.5, 1)$ are the coefficient of friction parameters for the legs of the table, $m=1$ is the mass of the box, $g=9.81$ is the gravitational acceleration, $\alpha = 4$ is the damping coefficient, and $\kappa = 100$ is the filter coefficient. The constraints \eqref{eq:model33} and \eqref{eq:model22} are exchanged with:
\begin{align*}
	& 0 \leq N_1 \perp -mg + N_1 + N_2 + N_3 \geq 0, \\
	& 0 \leq N_2 \perp -mg + N_1 + N_2 + N_3 \geq 0, \\
	& 0 \leq N_3 \perp -mg + N_1 + N_2 + N_3 \geq 0,
\end{align*}
to be consistent with the framework. \rev{ Note that extending the framework to LCS models with additional equality and inequality constraints as in \eqref{eq:model11}-\eqref{eq:model22} is straightforward but it is omitted for brevity.  }

\begin{figure}[t!]
	\hspace*{-0.5cm}
	\includegraphics[width=1.1\columnwidth]{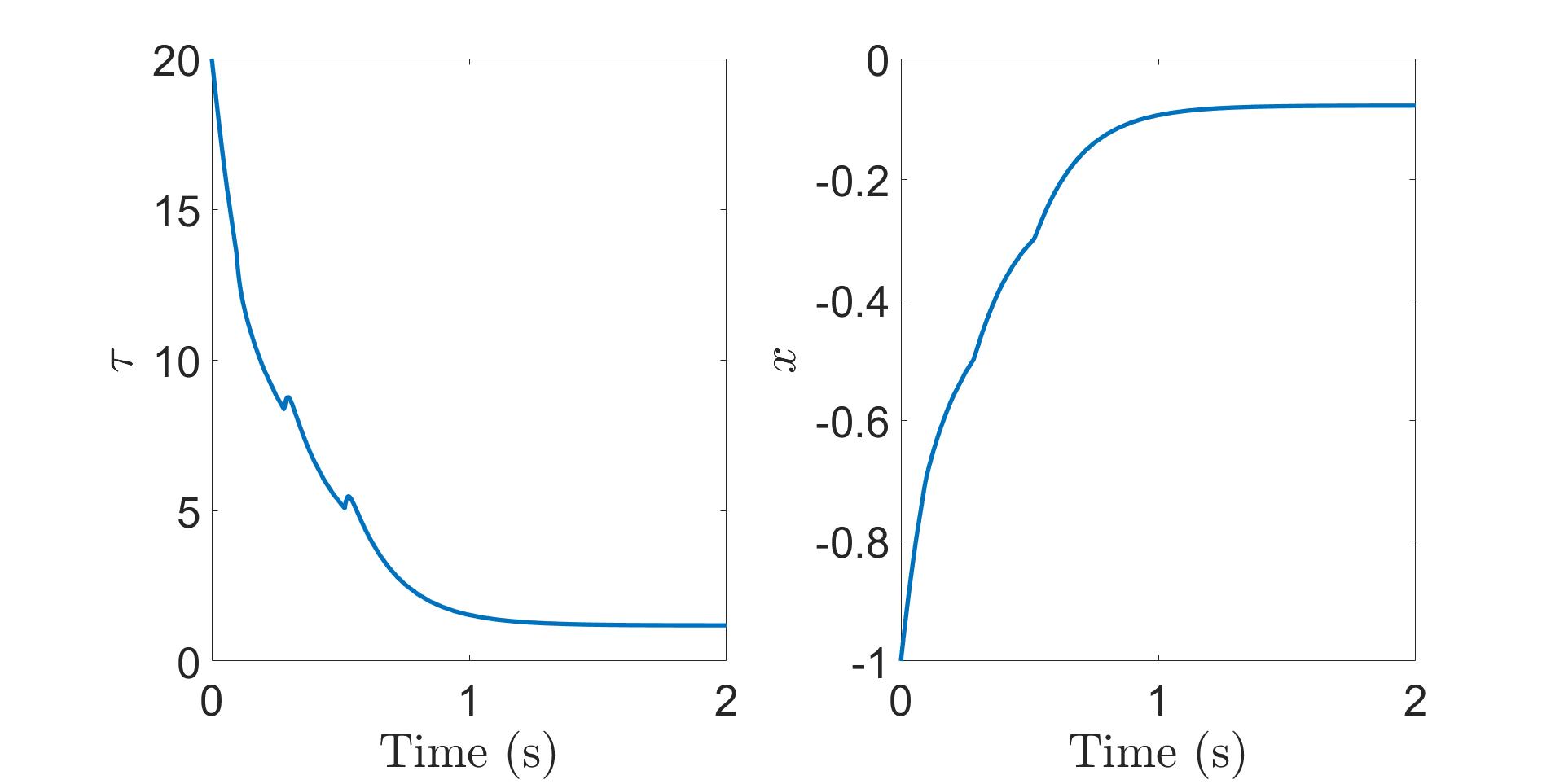}
	\caption{Simulation of three legged table example for the normal forces $N(t) = [4.0910, 4.1195, 1.5995]$ for $t \in [0, 0.2992)$, $N(t) = [5.4033, 3.1206, 1.2861]$ for $t \in [0.2992, 0.5455)$ and $N(t) = [9.4866, 0.1770, 0.1464]$ for $t \in [0.5455, \infty)$.}
	\label{fig:3legged_table}
\end{figure}

After using Algorithm 1, we find that $W = [0 \; 0 \; 0 \; 1 \; 1 \; 1]$ in \clr{242.8 minutes}. \clr{ Observe that one can solve a smaller sized polynomial optimization that only includes $N_1, N_2, N_3$ in 3.08 seconds as the variables are decoupled from $\gamma, \lambda_+, \lambda_-$ and reach the same result.} $W$ shows that $N_1 + N_2 + N_3$ is unique, as expected since $N_1 + N_2 + N_3 = mg$. Note that $W \lambda = mg$ is a constant and the Lyapunov function \eqref{eq:lyap_nonunique} reduces to a common Lyapunov function. Based on the structure of $W$, unlike the previous example, the net force $\lambda_+ - \lambda_-$ is not unique which is also expected due to the non-unique nature of normal forces. Notice that both the x-trajectory, $x(t)$ and $\lambda$-trajectory, $\lambda(t)$ are non-unique. \clr{The model has absolutely continuous solutions, $x(t)$ and $\tau(t)$,  for fixed $N_1,N_2,N_3$ following Proposition \ref{existence_uniqueness}. For this example, we only consider the case where $N_1(t), N_2(t), N_3(t)$ are bounded piecewise constant functions with finitely many pieces hence $x(t)$ and $\tau(t)$ are absolutely continuous.}

\rev{ The feasibility problem \eqref{eq:feasability_LCS} is solved in 19 seconds and a controller of the form $u(x,\lambda) = Kx + L \lambda$ is found such that the origin is Lyapunov stable } \clr{with $K = \begin{bmatrix}
	-20.75
	\end{bmatrix}$ and $L = \begin{bmatrix}
	0 & 0.36 & -0.36 & 0 & 0 & 0
	\end{bmatrix}$. The trajectories of the closed loop system are always absolutely continuous since we enforce that $L \text{SOL}(Ex+c,F)$ is a singleton for fixed, arbitrary $N_1$, $N_2$ and $N_3$.} 
In Figure \ref{fig:3legged_table}, observe that the force applied to the system ($\tau$) is continuous even though $u$ is not, due to the low-pass filter. The origin is Lyapunov stable and the trajectory does not reach origin because of stiction.

\begin{figure}[t]
	\hspace*{1.5cm}
	\includegraphics[width=0.7\columnwidth]{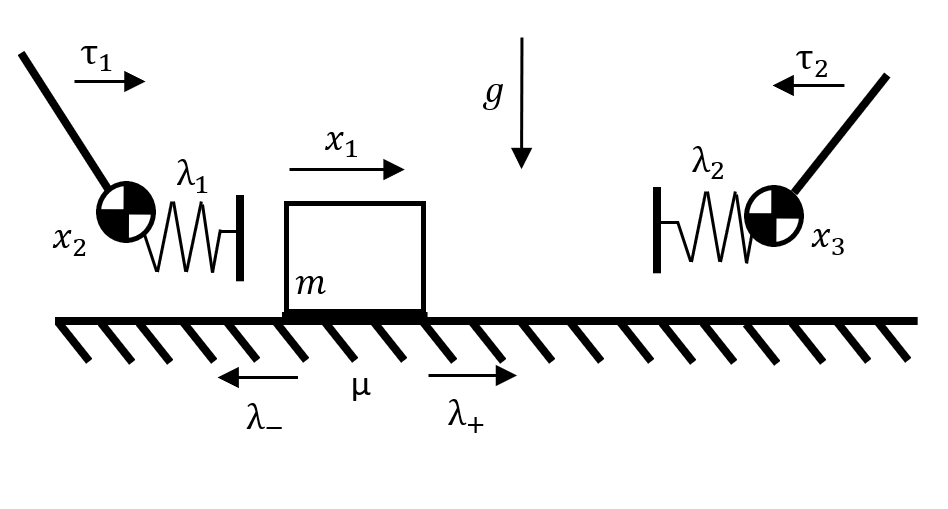}
	\caption{2D manipulation task where the goal is to regulate the position of the box on a surface with friction.}
	\label{fig:box_friction_manipulator}
\end{figure}

\subsection{2D Simple Manipulation}
\label{sub:friction3}
\rev{ Consider a quasi-static model of a box on a surface with friction parameter $\mu$ and two robotic arms that can interact with the box as in Figure \ref{fig:box_friction_manipulator}. } \rev{ Similar to the previous example, the force balance equation is used with Coulomb friction and damping to model the dynamics of the box. The velocity of the manipulators can be controlled directly with delayed inputs $\tau_1$ and $\tau_2$. } In this model $x_1, x_2, x_3$ represent the positions of the box, the left manipulator and the right manipulator respectively. The contact forces $\lambda_1$ and $\lambda_2$ are non-zero if and only if the distance between the manipulators and the box is less than some threshold. The friction force consists of a positive component $\lambda_+$ and a negative component $\lambda_-$. We assume that we can not observe anything related to the box except the contact force between the manipulators and the box. \rev{  The task is to regulate the box to the origin using the model: }
\begin{align*}
& \alpha \dot{x}_1 = \lambda_1 - \lambda_2 + \lambda_+ - \lambda_-, \\
& \dot{x}_2 = \tau_1, \\
& \dot{x}_3 = \tau_2, \\
& \dot{\tau}_1 = \kappa (u_1 - \tau_1), \\
& \dot{\tau}_2 = \kappa (u_2 - \tau_2), \\
& 0 \leq \lambda_1 \perp x_1 - x_2 + \frac{1}{k} \lambda_1 \geq 0, \\
& 0 \leq \lambda_2 \perp x_3 - x_1 + \frac{1}{k} \lambda_2 \geq 0, \\
& 0 \leq \gamma \perp \mu m g - \lambda_+ - \lambda_-  \geq 0, \\
& 0 \leq \lambda_+ \perp \gamma + \lambda_1 - \lambda_2 + \lambda_+ -\lambda_- \geq 0, \\
& 0 \leq \lambda_- \perp \gamma - \lambda_1 + \lambda_2 -\lambda_+ + \lambda_-  \geq 0,
\end{align*}
where $\kappa = 100$, $\mu = 0.1$, $m = 1$, $g = 9.81$, $\alpha = 1$, and $k=100$. Since, $F$ is not a P-matrix, we can construct the $W$ matrix using the result in Example \ref{sub:friction} and obtain
\begin{equation*}
	W = \begin{bmatrix}
	1 & 0 & 0 & 0 & 0 \\
	0 & 1 & 0 & 0 & 0 \\
	0 & 0 & 0 & 1 & -1 \\
	\end{bmatrix}.
\end{equation*}
Observing $W$, the net friction force, $\lambda_+ - \lambda_-$, is unique. The contact forces between the manipulators and the box, $\lambda_1, \lambda_2$ are also unique. \rev{Note that the closed-loop system has absolutely continuous solutions following Proposition \ref{existence_uniqueness} since it is enforced that $L \text{SOL}(Ex+c,F)$ is a singleton.}

\begin{figure}[t]
	\includegraphics[width=1.1\columnwidth]{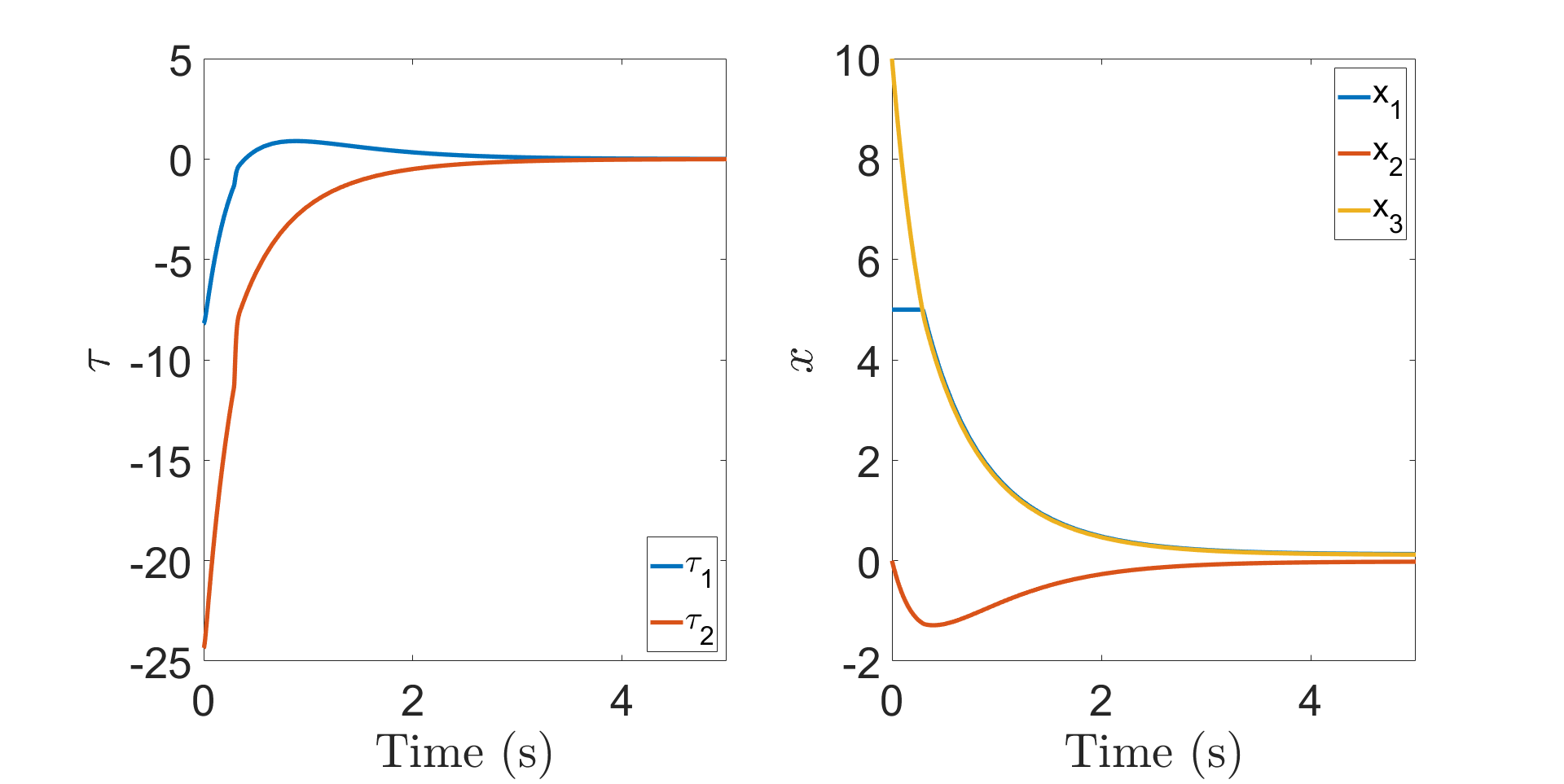}
	\caption{Simulation results for 2D simple manipulation example. The forces applied to the box ($\tau$) are smooth even though $u$ is not, due to the low-pass filter model.}
	\label{fig:manipulator_friction}
\end{figure}

Then we solve the optimization problem to find a controller that asymptotically stabilizes the system to a small ball around the origin $\mathcal{B} = \{ x : x^T x \leq 0.1 \}$. \rev{ The optimization problem finds a result in 7.06 minutes and a controller of the form $u(x,\lambda) = Kx + L\lambda$ that stabilizes the system is obtained \clr{with 
\begin{align*}
	& K = \begin{bmatrix}
	0 & -2.33 & -0.82 \\ 0 & -0.89 & -2.44
	\end{bmatrix},
	& L = \begin{bmatrix}
	-0.26 & 0.06 & 0 & 0 & 0 \\ -0.06 & 0.27 & 0 & 0 & 0
	\end{bmatrix}.
\end{align*}
} }
This example shows that the contact-aware policy can be used for systems with non-unique contact forces, e.g., quasi-static friction, where we do not have full state information. In Figure \ref{fig:manipulator_friction}, we demonstrate the performance of the controller.

\subsection{Four Carts}

\begin{figure}[b!]
	\hspace*{-0.2cm}
	\includegraphics[width=1\columnwidth]{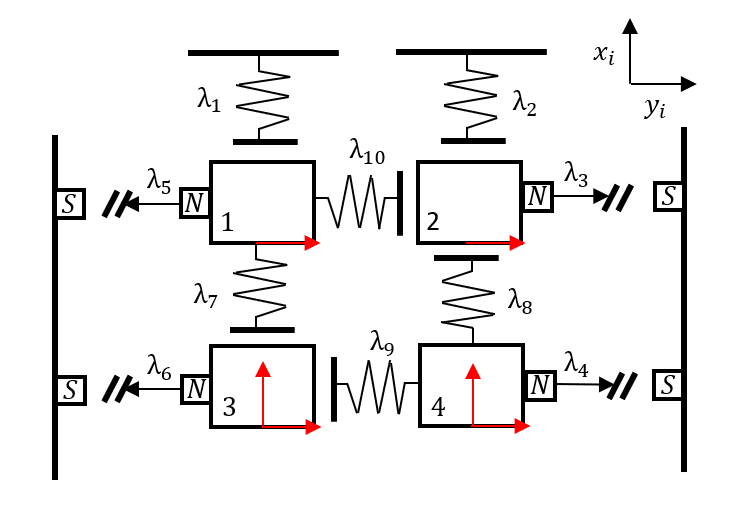}
	\caption{Four carts example. The inputs that are applied to carts are represented by the red arrows.}
	\label{fig: high_dim}
\end{figure}

\rev{ As our last example, consider the system in Figure \ref{fig: high_dim}. }
Here, $(x_i, y_i)$ gives the position of the cart $i$.
We approximate Newton's second law with a force balance equation for each cart.
The contact forces $\lambda_1, \lambda_2, \lambda_7, \lambda_8, \lambda_9$ and $\lambda_{10}$ are soft contacts that are represented by the springs and are non-zero if the objects are closer than a threshold. The forces $\lambda_3, \lambda_4, \lambda_5$ and $\lambda_6$ approximate attractive magnetic forces between the carts and the walls and similarly are non-zero if the distance between the carts and the walls is less than a threshold. The red arrows represent the input forces that can be applied to carts.
We model this system with $n=8$ states, and $m=10$ contacts where our goal is to show the performance of the proposed method on a high dimensional under-actuated example that is unstable without any control action.
The model parameters are $A = 0_{8 \times 8}$, $c = 0_{10 \times 1}$, $F = I_{10 \times 10}$,
\begin{equation*}
	B = \begin{bmatrix}
	\begin{pmatrix}
	1 & 0 \\
	0 & 0 \\
	0 & 1 \\
	0 & 0 \\
	\end{pmatrix} & 0_{4 \times 3} \\
	0_{3 \times 4} & I_{3 \times 3}
	\end{bmatrix},
\end{equation*}
\begin{equation*}
	D = \begin{bmatrix}
	0 & 0 & 0 & 0 & -1 & 0 & 0 & 0 & 0 & -1 \\
	-1 & 0 & 0 & 0 & 0 & 0 & 1 & 0 & 0 & 0 \\
	0 & 0 & 1 & 0 & 0 & 0 & 0 & 0 & 0 & 1 \\
	0 & -1 & 0 & 0 & 0 & 0 & 0 & 1 & 0 & 0 \\
	0 & 0 & 0 & 0 & 0 & -1 & 0 & 0 & -1 & 0 \\
	0 & 0 & 0 & 0 & 0 & 0 & -1 & 0 & 0 & 0 \\
	0 & 0 & 0 & 1 & 0 & 0 & 0 & 0 & 1 & 0 \\
	0 & 0 & 0 & 0 & 0 & 0 & 0 & -1 & 0 & 0 \\
	\end{bmatrix},
\end{equation*}
\begin{equation*}
E = \begin{bmatrix}
0 & -1 & 0 & 0 & 0 & 0 & 0 & 0 \\
0 & 0 & 0 & -1 & 0 & 0 & 0 & 0 \\
0 & 0 & -1 & 0 & 0 & 0 & 0 & 0 \\
0 & 0 & 0 & 0 & 0 & 0 & -1 & 0 \\
1 & 0 & 0 & 0 & 0 & 0 & 0 & 0 \\
0 & 0 & 0 & 0 & 1 & 0 & 0 & 0 \\
0 & 1 & 0 & 0 & 0 & -1 & 0 & 0 \\
0 & 0 & 0 & 1 & 0 & 0 & 0 & -1 \\
0 & 0 & 0 & 0 & -1 & 0 & 1 & 0 \\
-1 & 0 & 1 & 0 & 0 & 0 & 0 & 0 \\
\end{bmatrix}.
\end{equation*}

\begin{figure}[t!]
	\hspace*{-0.4cm}
	\includegraphics[width=1.1\columnwidth]{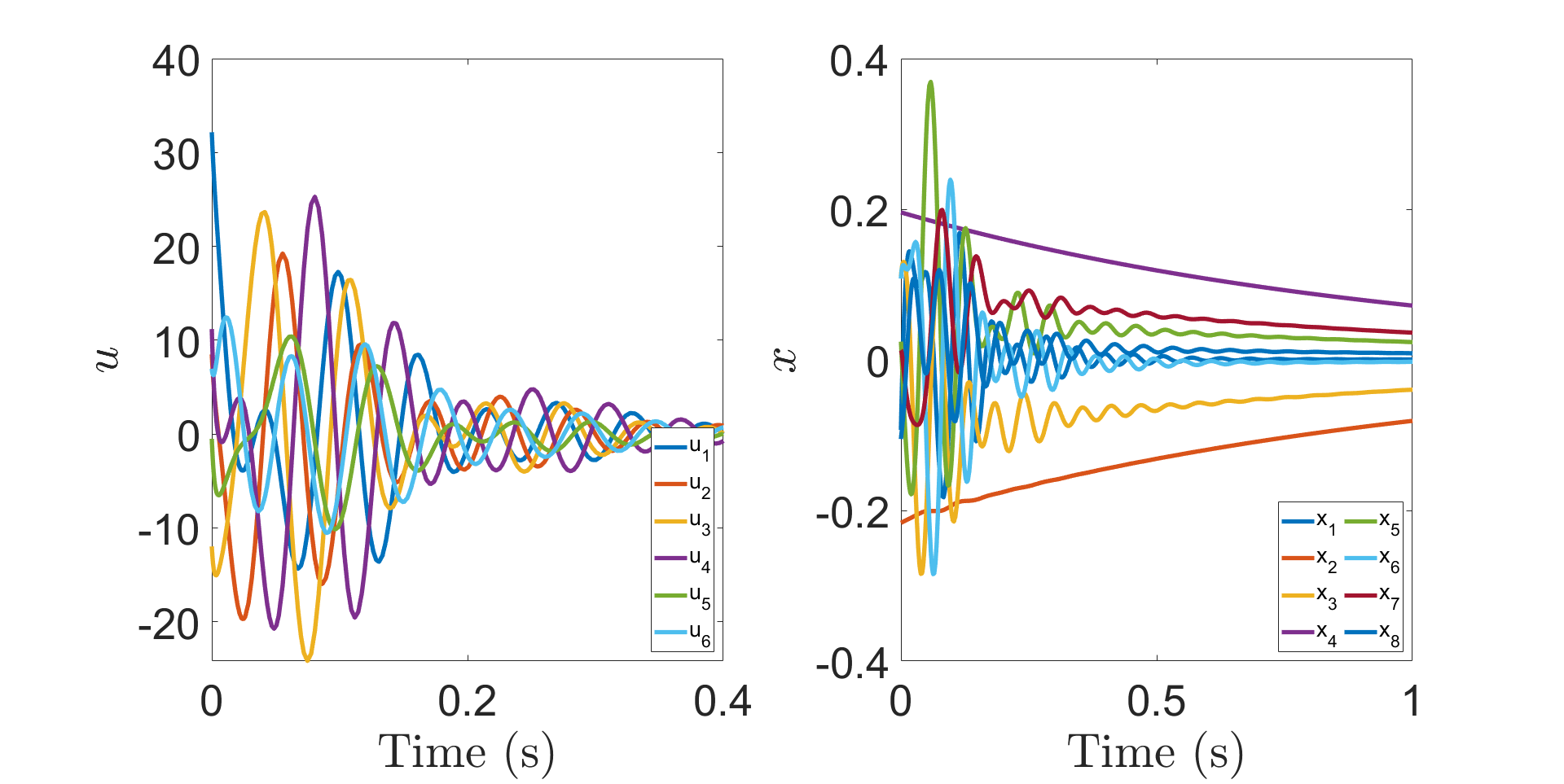}
	\caption{Simulation of four carts example. The state trajectory, $x(t)$, asymptotically converges to the origin. }
	\label{fig:four_carts}
\end{figure}

\clr{Observe that the model has absolutely continuous solutions following Proposition \ref{existence_uniqueness}.} \rev{A controller of the form $u(x,\lambda) = Kx + L \lambda$ 
that stabilizes the system can be found. }
For this example, and higher dimensional examples in general, the initialization of the $K$ and $L$ matrices have a significant affect on the success of the algorithm. 
We initialized elements of $K$ with sampling from uniform distribution ($K_{ij} \sim U[-100, 0]$).
For one successful case, the algorithm terminates in 6 minutes and 58 seconds, but we also needed to run the algorithm approximately 20 hours with random seeds to obtain a successful result. In Figure \ref{fig:four_carts}, we present the performance of the controller.

\begin{figure}[t!]
	\hspace*{-0.6cm}
	\includegraphics[width=1.1\columnwidth]{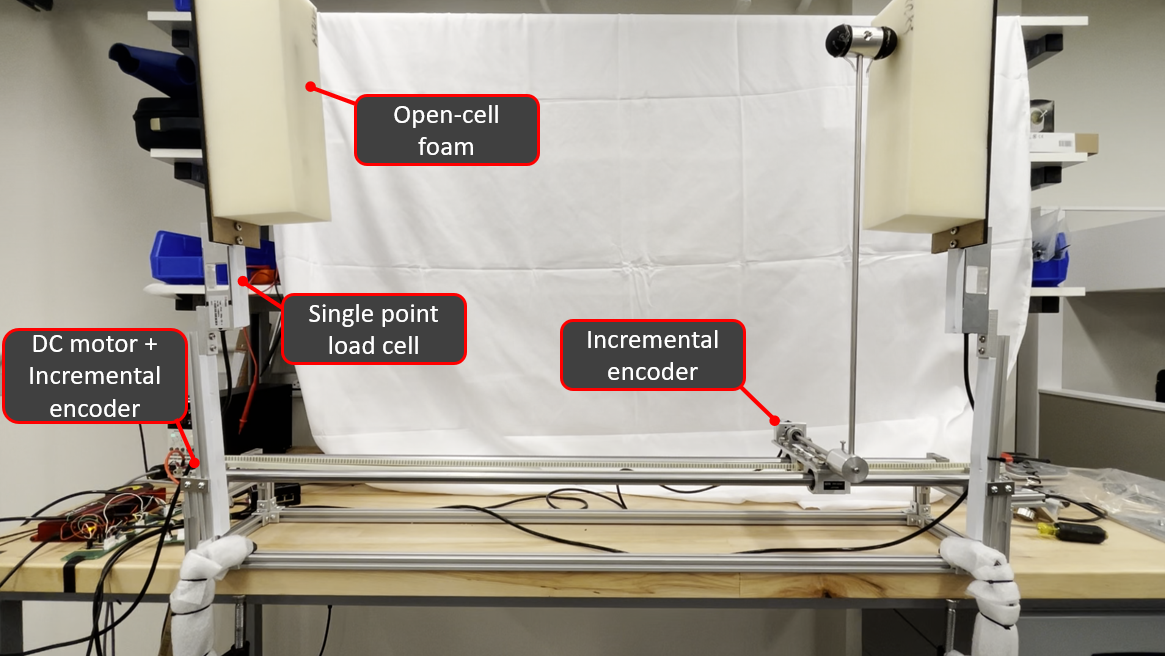}
	\caption{Experimental setup for cart-pole with soft walls.}
	\label{fig:cartpole_setup}
\end{figure}

\clr{
\section{Experimental Validation}
We demonstrate our contact-aware feedback controller on an experimental cart-pole system with soft walls shown in Figure \ref{fig:cartpole_setup}, replicating the system from Section \ref{subsection:cartpole}. \rev{ To generate linear motion of the cart, a DC motor is used with a belt drive. }
Linear motion of the cart was driven by torque-controlled DC motor and incremental encoders measured the positions of the cart and pole.
We added soft walls and tactile sensing capabilities; open-cell polyurethane foam was used for the walls, and a '\textit{Flintec PC42 Single Point Load Cell}' was used to measure the force exerted on the walls by the pole. Alternatively, sensors could have been placed on the pole itself.
}
\clr{
\subsection{System Model}
}
\clr{
We take the gravitational acceleration as $g = 9.81$, mass of the pole as $m_p = 0.35$, mass of the cart as $m_c = 0.978$, length of the center of mass location as $l_\text{CoM} = 0.4267$, the distance to the walls as $d=0.35$ and  and length of the pole as $l_p = 0.6$. 
After experimental trials, we identified the spring constant $k=700$ and neglect the damping term ($b=0$). The parameters of the LCS model are:
\begin{align*}
	& A = \begin{bmatrix}
	0 & 0 & 1 & 0 \\ 0 & 0 & 0 & 1 \\ 0 & 3.51 & 0 & 0 \\ 0 & 22.2 & 0 & 0
	\end{bmatrix}, \;
	B = \begin{bmatrix}
	0 \\ 0 \\ 1.02 \\ 1.7,
	\end{bmatrix} \\
	& D = \begin{bmatrix}
	0 & 0 \\ 0 & 0 \\ 0 & 0 \\ 4.7619 & -4.7619
	\end{bmatrix}, \;
	E = \begin{bmatrix}
	-1 & 0.6 & 0 & 0 \\ 1 & -0.6 & 0 & 0,
	\end{bmatrix}, \; \\
	& F = \begin{bmatrix}
	0.0014 & 0 \\ 0 & 0.0014
	\end{bmatrix}, \;
	c = \begin{bmatrix}
	0.35 \\ 0.35
	\end{bmatrix}.
\end{align*}
}
\begin{figure}[t!]
	\hspace*{-1.2cm}
	\includegraphics[width=1.3\columnwidth]{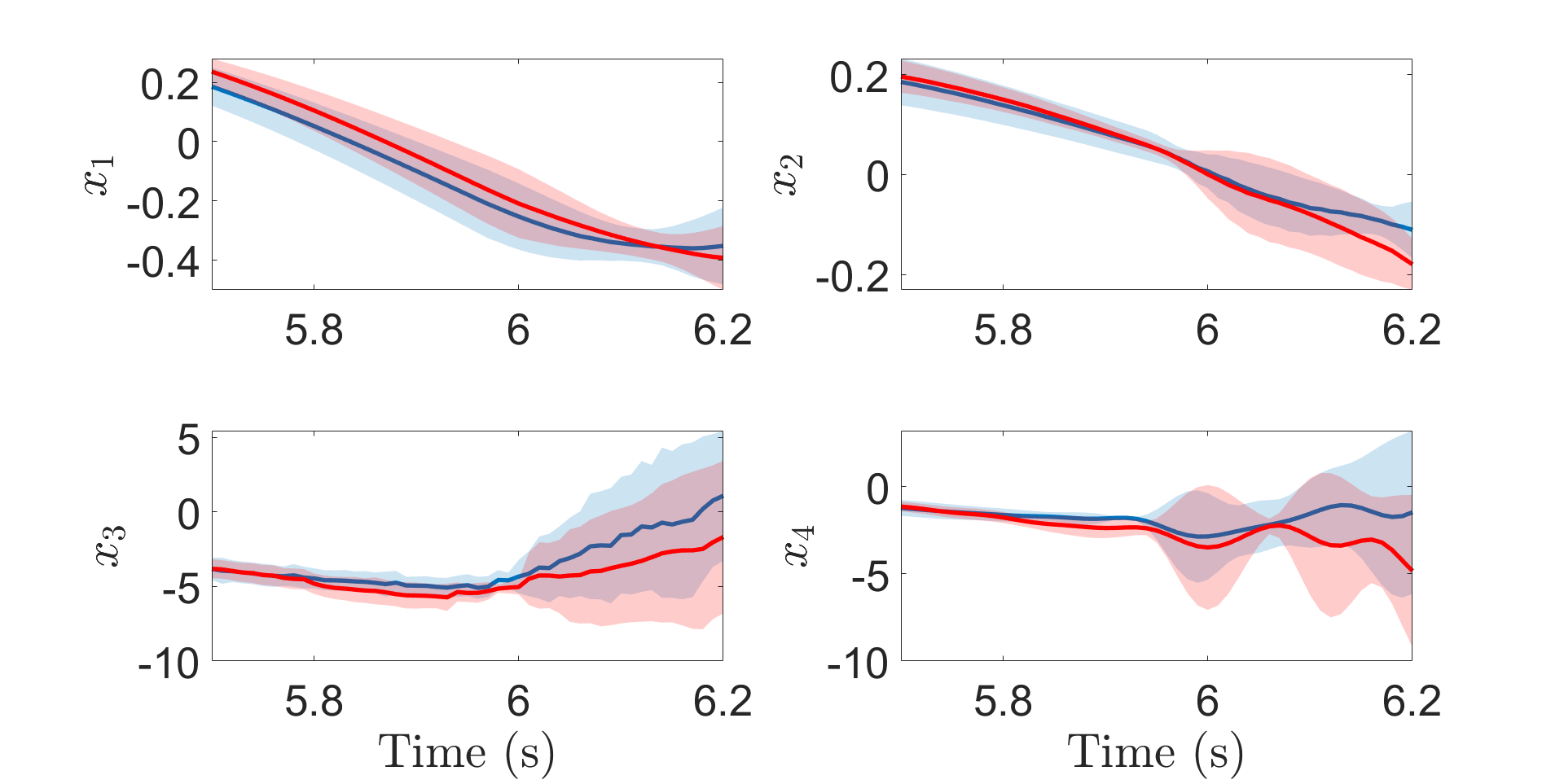}
	\caption{Experiment 1 - Trajectories around the impact event for all 6 trials. Blue represents contact-aware policy, red represents LQR, solid lines represent the respective means and shaded regions represent standard deviation.
		State distribution before the impact events are similar ($\approx 2$ cm difference in cart position and $\approx 1$ degree difference in pole angle). The velocity of the cart ($x_3$) is significantly higher for contact-aware policy after the impact.  }
	\label{fig:exp1_trialss}
\end{figure}
\begin{figure*}[t!]
	\includegraphics[width=2\columnwidth]{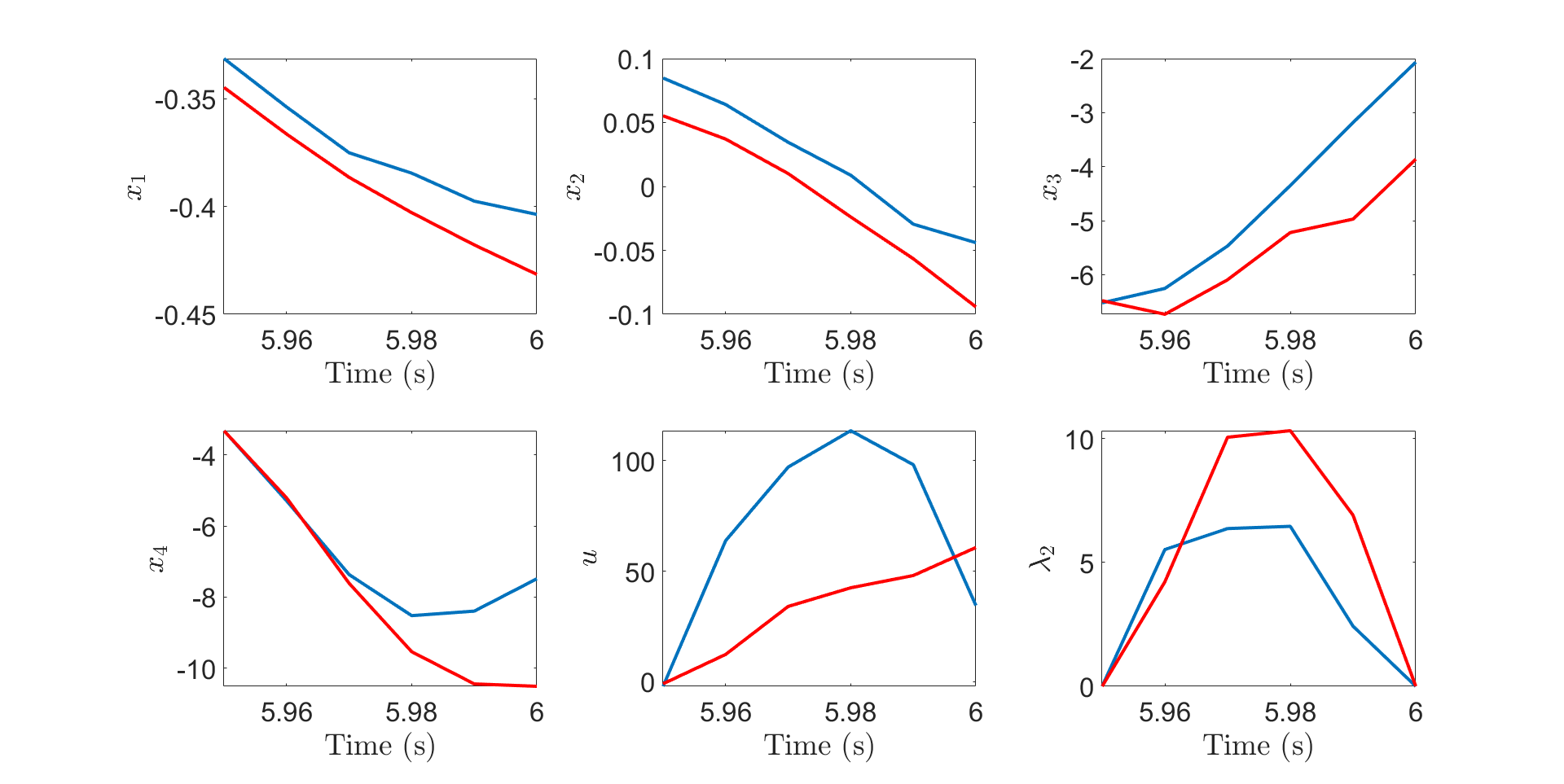}
	\caption{Experiment 1 - Blue represents the contact-aware policy and red represents LQR. Contact-aware policy ($u$) starts pushing the cart in the positive direction aggressively as soon as impact starts ($\lambda_2$) in order to catch the falling pole causing a big increase in the cart velocity ($x_3$). 
	As a result of contact-aware policy, the angular speed of pole is closer to zero ($x_4$). The contact-aware policy also mitigates the impact ($\lambda_2$). }
	\label{fig:exp1_specific}
\end{figure*}
\clr{
\subsection{Experiments}
Three sets of experiments are performed. For the first two, our method is compared against an LQR controller where $K_\text{LQR} = \begin{bmatrix}
3.16 & -40.78 & 4.3 & -7.67
\end{bmatrix}$. The contact-aware policy is designed such that $K = K_{\text{LQR}}$ is enforced and a contact gain matrix $L = \begin{bmatrix}
-10.02 & 10.02 
\end{bmatrix}$ is found.
It may be impossible to find a contact gain $L$ with a fixed $K$ in general, but $K = K_{\text{LQR}}$ was enforced for a more fair comparison with LQR. If we let the BMI design search for both $K$ and $L$, we can potentially get a better solution, though BMI enforces stability versus optimality.
Then, we perform between 6 and 10 trial experiments depending on the setup, after which we observed deterioration of the experimental setup due to the 
violence of the impact events that occurred when the LQR controller failed to stabilize the system. While performing these experiments, sensor readings below $1$ N are considered as $0$ N to neglect the effect of oscillations that occur after the impact events.
\rev{ For the third experiment, we repeat Experiment 1 without thresholds on the sensor readings. }
}
\newline

\paragraph{Experiment 1}
\clr{
We execute balancing controllers which attempt to stabilize the system to the origin and evaluate their performance by introducing large perturbations that lead to contact events.
First, the system is started in the upright position at the right wall, $x_0^T ~ = ~ \begin{bmatrix}
0.35 & 0 & 0 & 0
\end{bmatrix}$. Then, a control input\footnote{We apply the control input $u = K(x - x_s)$ where $x_s^T = \begin{bmatrix}
	0 & 0.35 & 0 & 0
	\end{bmatrix}$; hence, the cart moves towards the left wall; then, we switch back to $x_s^T = \begin{bmatrix}
	0 & 0 & 0 & 0
	\end{bmatrix}$.} 
is applied such that the pole impacts the left wall with high speed and close to upright position.
}

\clr{We repeated this experiment 6 times for both policies. In Figure \ref{fig:exp1_trialss}, we demonstrate that the initial conditions for all trials (at $t=5.7$) are similar. 
The mean difference is $0.05$, $0.01$, $0.01$, $0.04$ respectively for $x_1$, $x_2$, $x_3$, $x_4$.
The LQR policy failed in all ($0/6$) of the trials whereas contact-aware policy was always successful ($6/6$).
Since the policies are identical when not in contact, we focus our analysis and plots on the brief time windows ($t = [5.7, 6.2]$ as in Figure \ref{fig:exp1_trialss}) which contain impact events.
The contact-aware policy results in a significant increase in cart velocity (during the impact event) in order to catch the falling pole (the mean cart velocity for contact-aware policy is $2.76$ m/s higher than LQR at $t=6.2$). Similarly, the mean angular velocity of the pole with LQR controller is $-4.84$ rad/s compared to $-1.48$ rad/s of contact-aware policy which also demonstrates that contact-aware policy is reacting better to the falling pole as expected.
}

\clr{
We examine a specific trial (Figure \ref{fig:exp1_specific}) where the differences between the pre-impact states are $0.01$, $0.02$, $0.2$ and $0.01$ for $x_1, x_2, x_3, x_4$ respectively. Over the $50$ ms impact period, change in the cart velocity with contact-aware controller is $2.2$ m/s higher than LQR. Similarly, the change in angular velocity of the pole is $3.01$ rad/s more than LQR. As shown in Figure \ref{fig:exp1_specific}, as soon as the impact event with the left wall starts the contact-aware policy tries to push the cart in the positive direction in order to catch the falling pole and stabilizes the system unlike LQR.
}

\begin{figure}[b!]
	\hspace*{-0.8cm}
	\includegraphics[width=1.1\columnwidth]{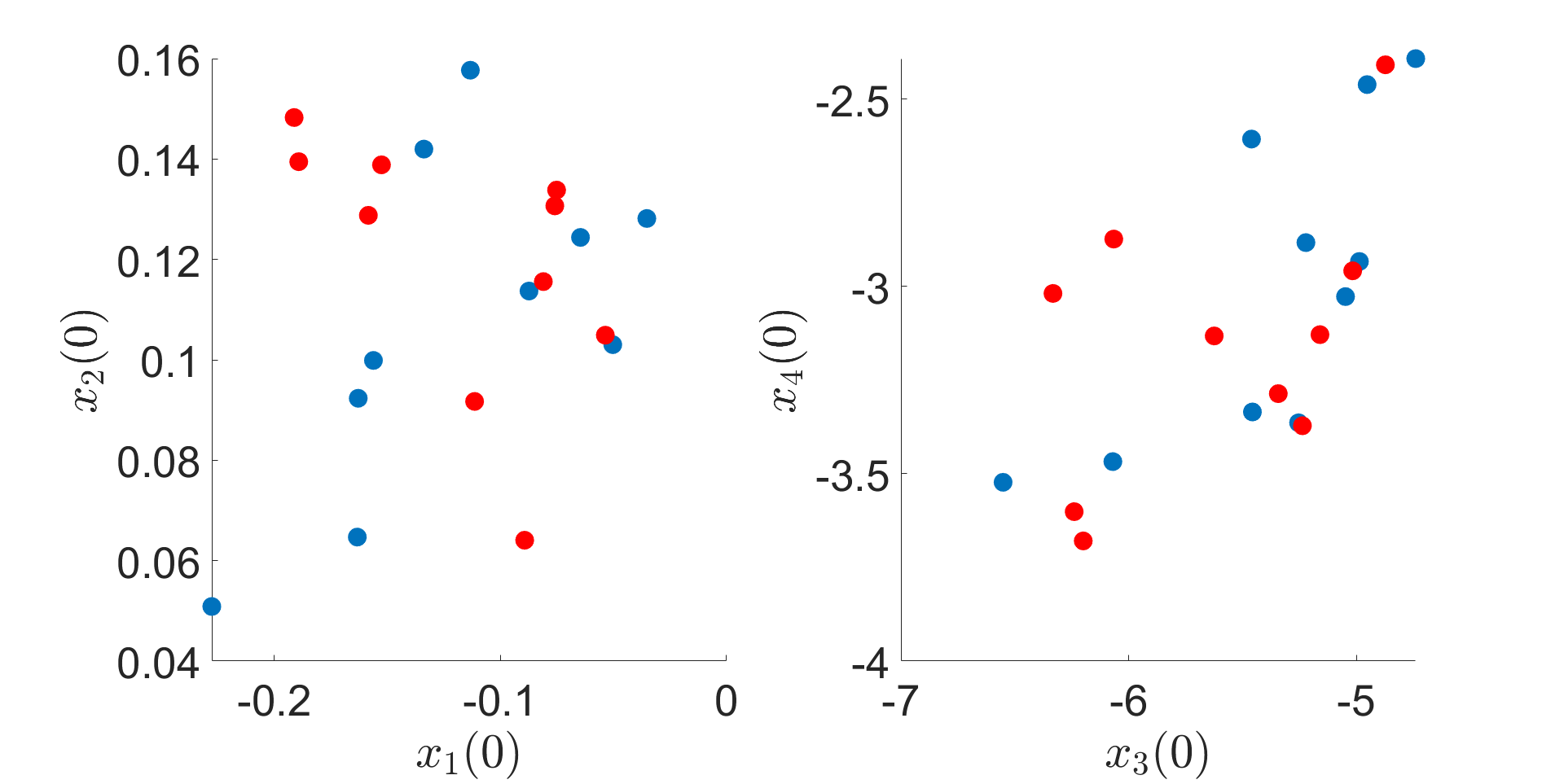}
	\caption{Experiment 2 - Distribution of initial conditions where blue represents contact-aware policy trials and red represents LQR trials.}
	\label{fig:exp2_distribution}
\end{figure}

\paragraph{Experiment 2}
\clr{
In the first experiment, we created a consistent initial condition across all trials. Here, we introduce random perturbations to cover a broader range of initial conditions. As with the first experiment, the goal of the initialization process is to create conditions which initiate contact.
First, the system is balanced at the origin. 
Then, we apply a control input\footnote{We apply the control input $u = K(x - x_s)$ where
 $x_s^T = \begin{bmatrix}
	0 & 0.5 & 0 & 0 
\end{bmatrix}$ for 0.5 seconds and switch back to $x_s^T = \begin{bmatrix}
0 & 0 & 0 & 0 
\end{bmatrix}$.} 
briefly to ensure that pole is close to impacting the left wall with a relatively high speed. Next, we apply a random input disturbance with uniform distribution $u_d \sim U[5, 10]$ for $100$ ms. We repeat this experiment 10 times for each LQR and contact-aware policy (with same seeds). After the random input disturbance is applied, the states are distributed as shown in Figure \ref{fig:exp2_distribution}. 
}

\begin{figure}[t!]
	\hspace*{-0.8cm}
	\includegraphics[width=1.2\columnwidth]{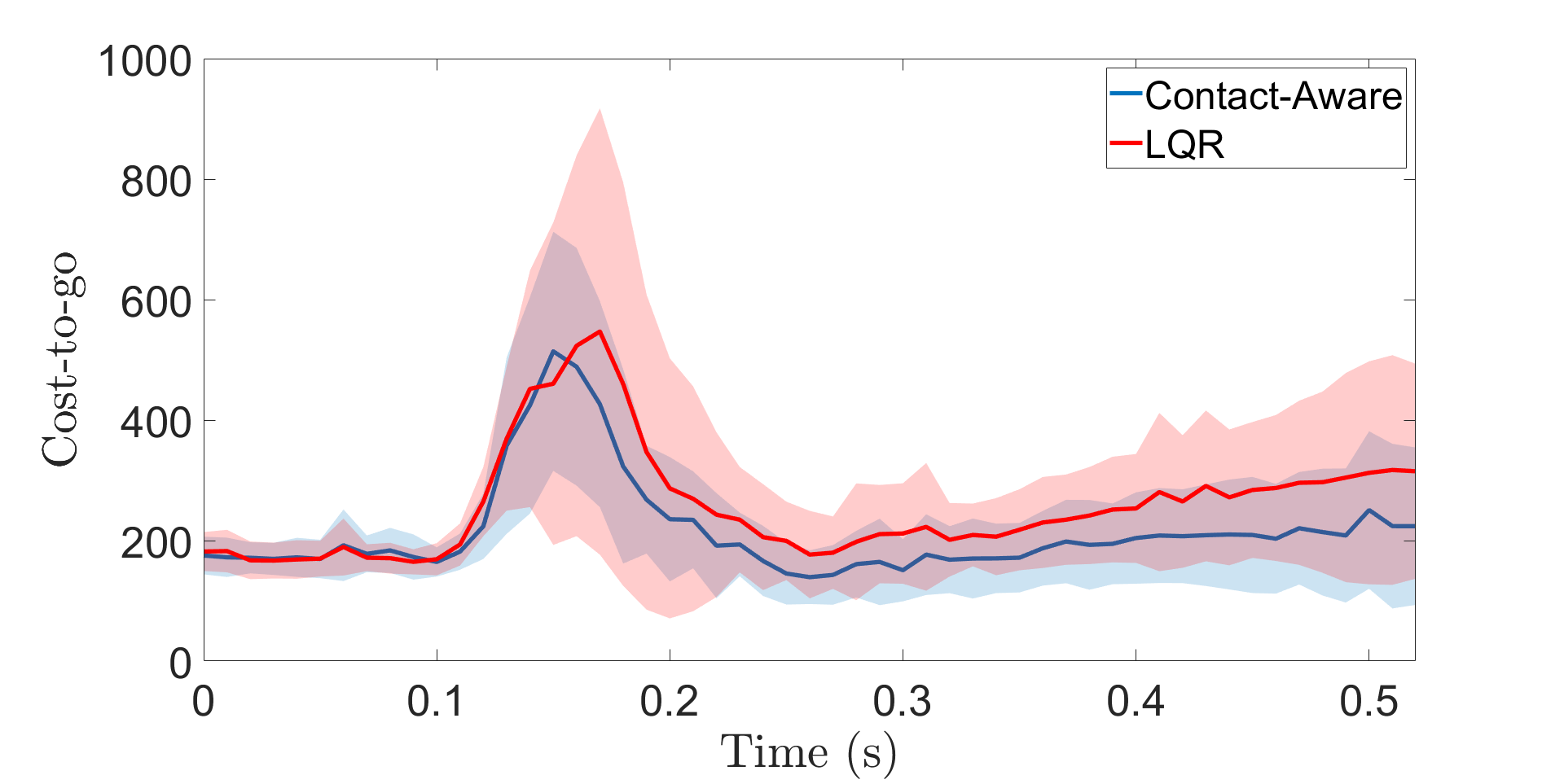}
	\caption{Experiment 2 - LQR cost-to-go for all trials during the impact event (impact events are aligned for all trials). Blue represents contact-aware policy, red represents LQR, solid lines represent the respective means and shaded regions represent standard deviation. Contact-aware cost-to-go surpasses LQR cost-to-go as the impact starts due to the aggressive tactile feedback but contact-aware policy ends up with a lower cost-to-go after the impact event. }
	\label{fig:exp2_cost_impact}
\end{figure}

\begin{figure}[b!]
	\hspace*{-1cm}
	\includegraphics[width=1.2\columnwidth]{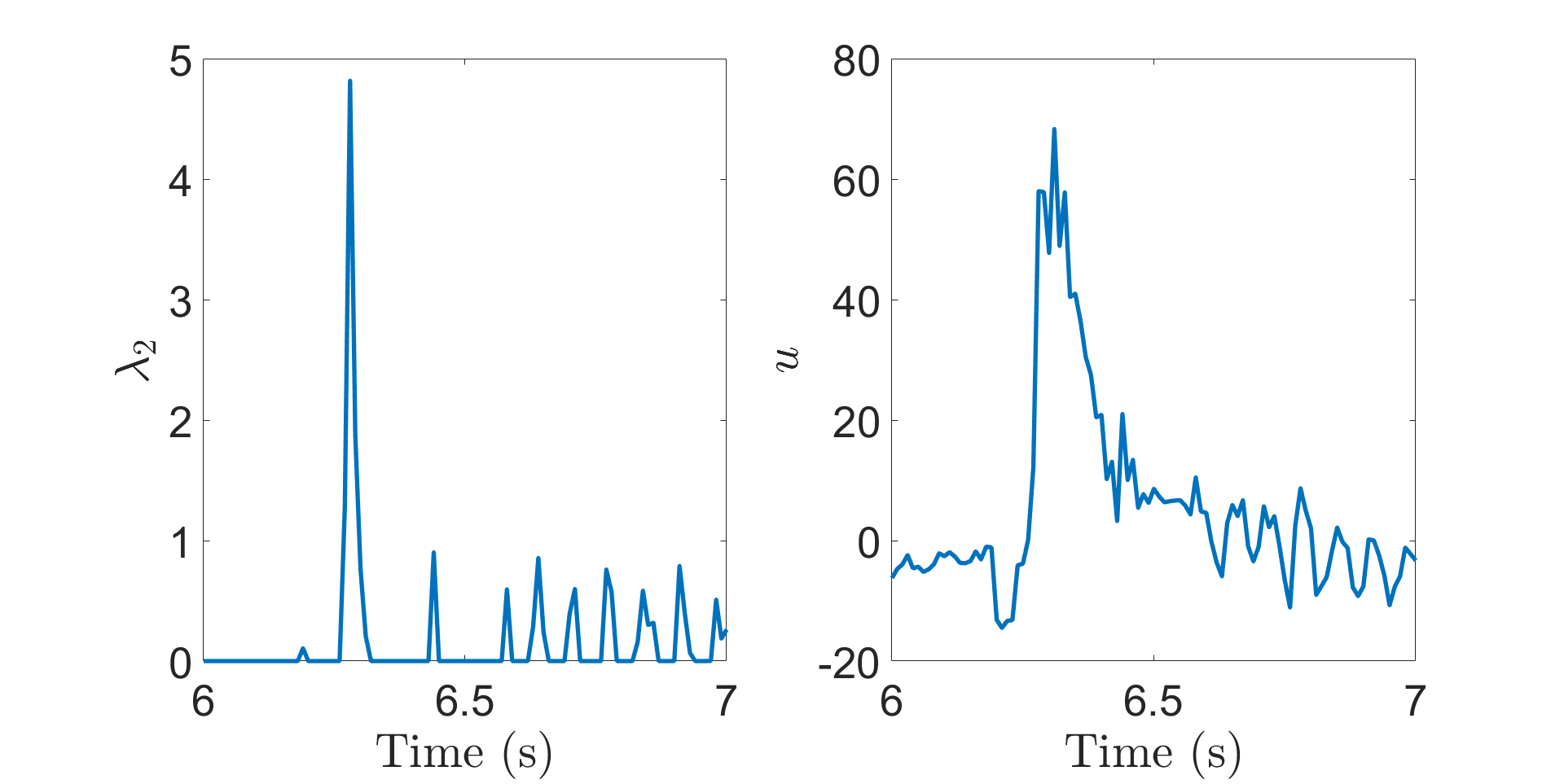}
	\caption{Experiment 3 - The oscillations in sensor readings ($\lambda_2$) that are caused by the impact event and the corresponding control action
		without heuristic-based thresholds. }
	\label{fig:no_cutoff}
\end{figure}

\clr{Out of the 10 trials, the LQR controller failed in 5/10 of the trials whereas the contact-aware policy was always successful. In Figure \ref{fig:exp2_cost_impact}, we demonstrate that the contact-aware policy ends up with a lower cost-to-go than LQR after the impact event. \rev{ Note that LQR cost-to-go is a useful metric, more so than the 2-norm, since it represents an approximate cost to complete the stabilization task.}
}

\begin{figure}[t!]
	\includegraphics[width=1\columnwidth]{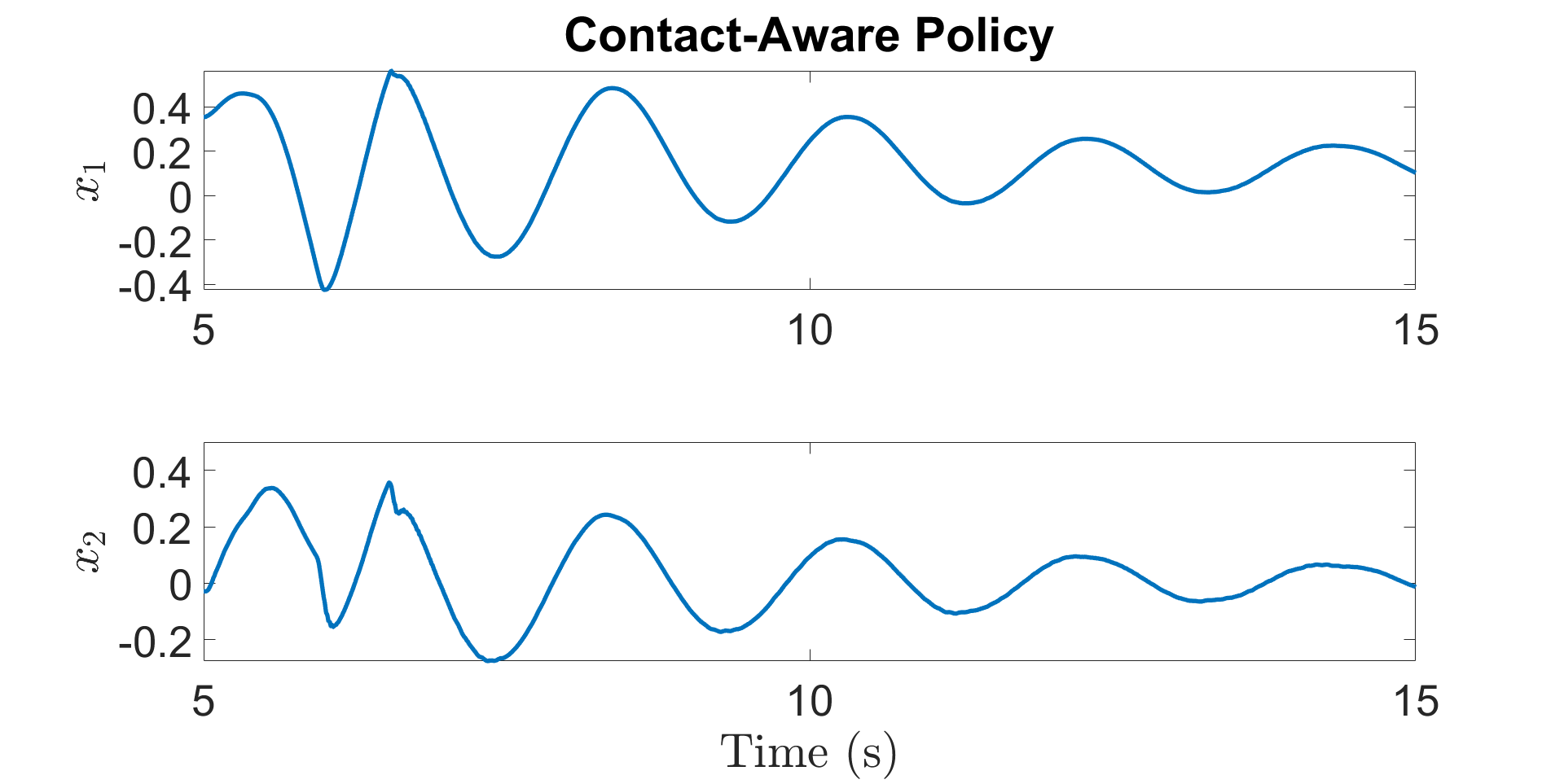}
	\caption{Experiment 3- Trajectory with contact-aware controller without any heuristic-based thresholds.}
	\label{fig:exp3_traj}
\end{figure}

\begin{figure}[b!]
	\includegraphics[width=1\columnwidth]{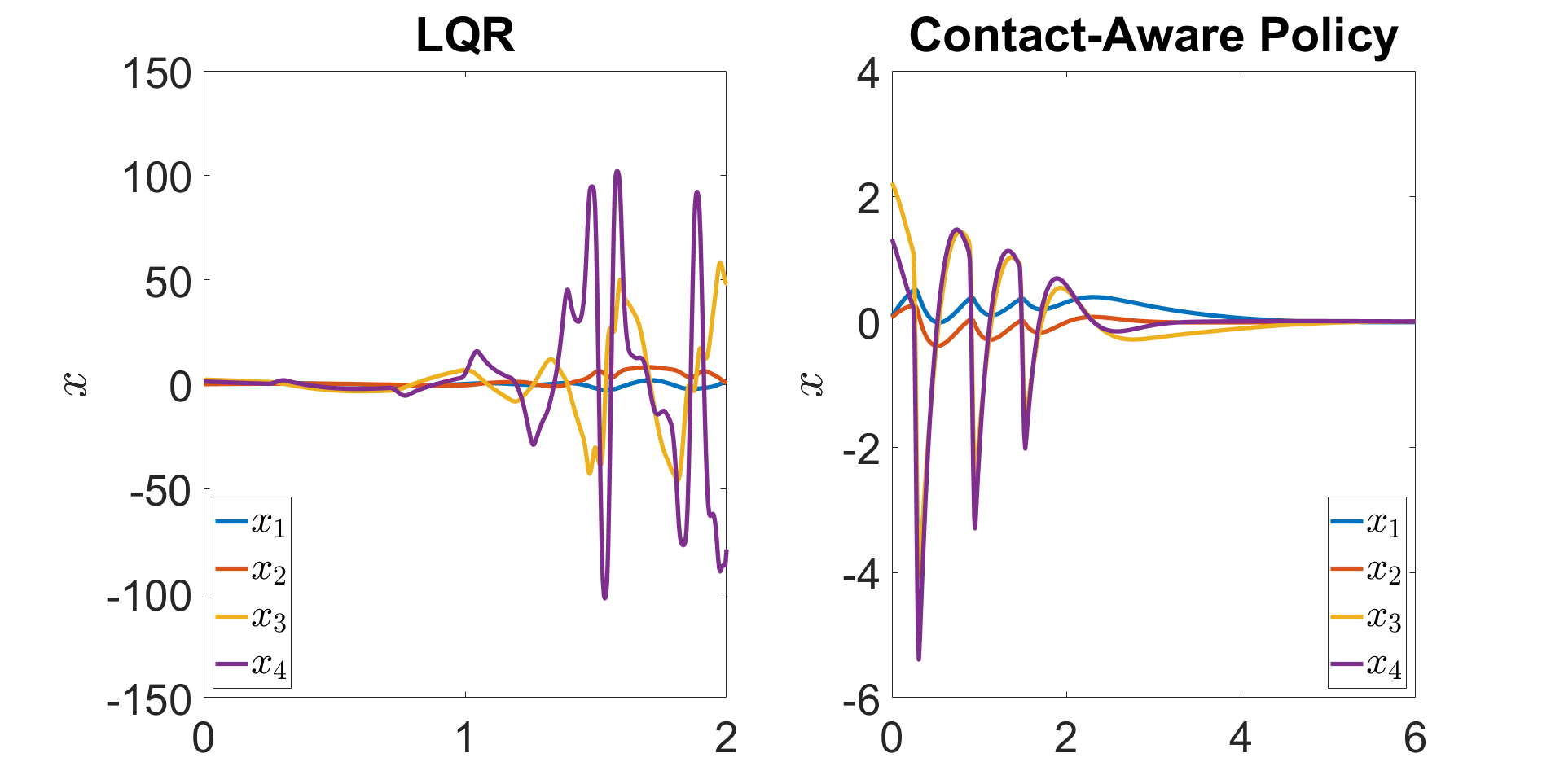}
	\caption{Experiment 3- Simulation results (as in Section \ref{subsection:cartpole}) where we simulate forward from a state obtained from the experiment. Simulation captures the system response qualitatively as LQR is unstable and contact-aware policy is successful.}
	\label{fig:exp3_sim_traj}
\end{figure}

\paragraph{Experiment 3}
\clr{ After the impact event, the walls oscillate back and forth which causes oscillations in sensor readings as shown in Figure \ref{fig:no_cutoff}.
In this experiment, we do not apply a threshold to the sensor readings. This enables pure feedback on sensor measurements, rather than heuristic-based thresholds or slow/inaccurate mode detection (where the state-of-the-art takes 4-5 ms \cite{bledt2018cheetah}) that purely hybrid approaches utilize. Hence, we repeat the procedure in Experiment 1 and observe that the controller is successful in stabilizing the system (Figure \ref{fig:exp3_traj}) even  without heuristic-based thresholds.
\rev{In Figure \ref{fig:exp3_sim_traj}, model is simulated (as in Section \ref{subsection:cartpole}) starting from the an initial condition obtained from the experiment and contact-aware policy stabilizes the system whereas LQR fails. This demonstrates that simulations capture the system qualitatively.}
}

\section{Conclusion}
In this work, we have introduced a controller that can utilize both state and force feedback.
We have demonstrated that combining linear complementarity systems with such tactile feedback controllers might result in an algebraic loop, and discussed how one can break such algebraic loops.

We have proposed an algorithm for synthesizing contact-aware control policies for linear complementarity systems with possibly non-unique solutions.
For soft contact models, we have shown that pure local, linear analysis was entirely insufficient and utilizing contact in the control design is critical to achieve high performance. For systems with non-unique solutions, we have proposed a polynomial optimization program that can find matrices that map non-unique contact forces into a unique value, and used such mappings in our controller design algorithm. We have shown the effectiveness of our method on quasi-static friction models.

Furthermore, the proposed algorithm exploits the complementarity structure of the system and avoids enumerating the exponential number of potential modes, enabling efficient design of multi-contact control policies. 
Towards this direction, we have presented an example with eight states and ten contacts.
In addition to incorporating tactile sensing into dynamic feedback, we provide stability guarantees for our design method \clr{and we have verified our method on an experimental setup}.

The algorithm requires solving feasibility problems that include bilinear matrix inequalities and we have used PENBMI \cite{kovcvara2003pennon}. For the examples presented here, except the last one, the runtime of the algorithm was short and we found solutions to the problems relatively quickly. On the other hand, it is important to note that for some parameter choices and initializations, the solver was unable to produce feasible solutions.

Interesting future work in this area will be using the controller presented here in physical experiments. We consider a hierarchical control framework where the tactile feedback policy is the higher level controller working together with a lower level controller to achieve a specified task. In addition, we intend to extend these algorithms to more complex tasks. For example, quasi-static models \cite{halm2019quasi} where the matrix $F$ depends on the generalized coordinates $q$. Another direction is designing controllers for systems where there are bilinear terms $(x_i \lambda_j)$ in the dynamics, since we believe that bilinear terms are important when locally approximating a certain class of non-smooth systems. \clr{Also, works such as \cite{brogliato1999nonsmooth} draw connections between compliant and rigid contact models. Such approaches can help analyzing the controllers designed in this work for a range wider range of models.}
\rev{Lastly, it may be possible to increase the application of our method by utilizing non-monotonic and almost-decreasing Lyapunov functions \cite{moruarescu2010trajectory}.}



\section*{APPENDIX A}

\clr{Next, we present the matrix inequalities in \eqref{eq:ineq1} and \eqref{eq:ineq2} explicitly. We can represent \eqref{eq:ineq1} as:
	\begin{align*}
	&T_1 - S_1^T W_1 S_1 - \frac{1}{2} ( S_{2,1} + S_{2,1}^T ) \succeq 0, \\
	&T_2 + S_1^T W_2 S_1 + \frac{1}{2} ( S_{2,2} + S_{2,2}^T ) \preceq 0,
	\end{align*}
	where $W_i$ are decision variables with non-negative entries, $J_i = \text{diag}(\tau_i)$ where $\tau_i$ are free decision variables, and
	\begin{align*}
	&T_1 = \begin{bmatrix} P - \gamma_1 I & Q & p/2 \\ * & R & r/2 \\ * & * & z \end{bmatrix}, 
	T_2 = \begin{bmatrix} P - \gamma_2 I & Q & p/2 \\ * & R & r/2 \\ * & * & z \end{bmatrix}, \\
	&S_1 = \begin{bmatrix} E & F & c \\ 0 & I & 0 \\ 0 & 0 & 1 \end{bmatrix},
	S_{2,i} = \begin{bmatrix} 0 & 0 & 0 \\ J_i E & J_i F & J_i c \\ 0 & 0 & 0 \end{bmatrix}.
	\end{align*}
	We can represent \eqref{eq:ineq1} as:
	\begin{align*}
	T_3 + S_3^T W_3 S_3 + \frac{1}{2} ( S_4 + S_4^T ) + \sum_{i=1}^m \frac{1}{2} (S_{5,i} + S_{5,i}^T) \preceq 0,
	\end{align*}
	where $W_i$ are decision variables with non-negative entries, $J_i = \text{diag}(\tau_i)$ where $\tau_i$ are free decision variables and $\zeta_{5,i} = \text{diag}(\theta_i)$ where $\theta_i$ are zero everywhere expect the $i$th entry which is a free variable and ($T_3$ is symmetric)
	\begin{align*}
	\footnotesize
	T_3 = \begin{bmatrix} PA + A^T P & PD + A^T Q & Q & A^T p/2 + P a & 0 & 0 \\
	* & D^T Q + Q^T D & R & Q^T a + D^T p / 2 & 0 & 0 \\
	* & * & 0 & r/2 & 0 & 0\\
	* & * & * & p^T a & 0 & 0 \\
	0 & 0 & 0 & 0 & 0 & 0 \\
	0 & 0 & 0 & 0 & 0 & 0
	\end{bmatrix},
	\end{align*}
	\begin{align*}
	&S_3 = \begin{bmatrix} E & F & 0 & c & 0 & 0 \\ 
	0 & I & 0 & 0 & 0 & 0 \\ 
	0 & 0 & 0 & 1 & 0 & 0,
	\end{bmatrix}, \\
	&S_4 = \begin{bmatrix} 0 & 0 & 0 & 0 & 0 & 0 \\
	J_3 E & J_3 F & 0 & J_3 c & 0 & 0 \\
	0 & 0 & 0 & 0 & 0 & 0\\
	J_4 EA & J_4 ED & J_4 F + J_5 & J_4 E a & J_4 & J_5 \\
	0 & 0 & 0 & 0 & 0 & 0 \\
	0 & 0 & 0 & 0 & 0 & 0
	\end{bmatrix},\\
	&S_{5,i} = \begin{bmatrix} 0 & 0 & 0 & 0 & 0 & 0 \\
	0 & 0 & 0 & 0 & \zeta_{7,i} & 0 \\
	0 & 0 & 0 & 0 & 0 & 0\\
	0 & 0 & 0 & 0 & 0 & 0 \\
	0 & 0 & 0 & 0 & 0 & \zeta_{8,i} \\
	\zeta_{9,i} E & \zeta_{9,i} F & 0 & \zeta_{9,i} c & 0 & 0
	\end{bmatrix}.
	\end{align*}
}

\section*{APPENDIX B}
\clr{
We present the two polynomial optimization problems. The first one is regarding Proposition \ref{unique_map}:
\begin{alignat}{2}
\label{eq:find_W_poly_exp}
& \underset{}{\text{find}} && w, \eta, p^k_{i}, p^k_{i,j}, s^k_i \\
\notag & \text{subject to}  \quad && \phi_1(\boldsymbol{\lambda}_1, \boldsymbol{\lambda}_2, \boldsymbol{q}) \geq 0, \\
\notag & &&  \phi_2(\boldsymbol{\lambda}_1, \boldsymbol{\lambda}_2, \boldsymbol{q}) \geq 0,
\end{alignat}
and the second one considers the optimization problem in \eqref{eq:find_W_poly}:
\begin{alignat}{2}
\label{eq:find_W_poly_explicit}
& \underset{w, \eta, p^k_i, p^k_{i,j} , s^k_i  }{\text{min}} && r^T N^T w \\
\notag& \text{subject to}  \quad && \phi_1(\boldsymbol{\lambda}_1, \boldsymbol{\lambda}_2, \boldsymbol{q}) \geq 0, \\
\notag& &&  \phi_2(\boldsymbol{\lambda}_1, \boldsymbol{\lambda}_2, \boldsymbol{q}) \geq 0,  \\
\notag& && |w_i| \leq 1, \; \forall i, \; \eta \geq 0,
\end{alignat}
where the functions $\phi_1$ and $\phi_2$ are defined as
\begin{align*}
	\phi_1(\boldsymbol{\lambda}_1, \boldsymbol{\lambda}_2, \boldsymbol{q}) &= (\eta + w^T (\boldsymbol{\lambda}_1 - \boldsymbol{\lambda}_2)) (\boldsymbol{\lambda}_1^T \boldsymbol{\lambda}_1 + \boldsymbol{\lambda}_2^T \boldsymbol{\lambda}_2) \\
	& + \sum_i p^1_i \boldsymbol{\lambda}_{1,i} + \sum_i p^2_i \boldsymbol{\lambda}_{2,i} + \sum_i \sum_j p^3_{i,j} \boldsymbol{\lambda}_{1,i} \boldsymbol{\lambda}_{1,j} \\
	& + \sum_i p^4_i (\boldsymbol{q}_i + F_i^T \boldsymbol{\lambda}_1) + \sum_i p^5_i (\boldsymbol{q}_i + F_i^T \boldsymbol{\lambda}_2) \\
	& + \sum_i \sum_j p^6_{i,j} (\boldsymbol{q}_i + F_i^T \boldsymbol{\lambda}_1) (\boldsymbol{q}_j + F_j^T \boldsymbol{\lambda}_2) \\
	& + \sum_i \sum_j p^7_{i,j} (\boldsymbol{q}_i + F_i^T \boldsymbol{\lambda}_2) \boldsymbol{\lambda}_{1,j} \\
	& + \sum_i \sum_j p^8_{i,j} (\boldsymbol{q}_i + F_i^T \boldsymbol{\lambda}_1) \boldsymbol{\lambda}_{2,j}\\
	& + \sum_i s^1_i \boldsymbol{\lambda}_{1,i} (\boldsymbol{q}_i + F_i^T \boldsymbol{\lambda}_1) \\
	& + \sum_i s^2_i \boldsymbol{\lambda}_{2,i} (\boldsymbol{q}_i + F_i^T \boldsymbol{\lambda}_2),
\end{align*}
where $p^k_{i}, p^k_{i,j}$ are non-negative variables (sum-of-squares polynomials), $ s^k_i$ are free variables (polynomials with no restriction) and
\begin{align*}
\phi_2(\boldsymbol{\lambda}_1, \boldsymbol{\lambda}_2, \boldsymbol{q}) &= (\eta - w^T (\boldsymbol{\lambda}_1 - \boldsymbol{\lambda}_2)) (\boldsymbol{\lambda}_1^T \boldsymbol{\lambda}_1 + \boldsymbol{\lambda}_2^T \boldsymbol{\lambda}_2) \\
& + \sum_i p^9_i \boldsymbol{\lambda}_{1,i} + \sum_i p^{10}_i \boldsymbol{\lambda}_{2,i} + \sum_i \sum_j p^{11}_{i,j} \boldsymbol{\lambda}_{1,i} \boldsymbol{\lambda}_{1,j} \\
& + \sum_i p^{12}_i (\boldsymbol{q}_i + F_i^T \boldsymbol{\lambda}_1) + \sum_i p^{13}_i (\boldsymbol{q}_i + F_i^T \boldsymbol{\lambda}_2) \\
& + \sum_i \sum_j p^{14}_{i,j} (\boldsymbol{q}_i + F_i^T \boldsymbol{\lambda}_1) (\boldsymbol{q}_j + F_j^T \boldsymbol{\lambda}_2) \\
& + \sum_i \sum_j p^{15}_{i,j} (\boldsymbol{q}_i + F_i^T \boldsymbol{\lambda}_2) \boldsymbol{\lambda}_{1,j} \\
& + \sum_i \sum_j p^{16}_{i,j} (\boldsymbol{q}_i + F_i^T \boldsymbol{\lambda}_1) \boldsymbol{\lambda}_{2,j}\\
& + \sum_i s^3_i \boldsymbol{\lambda}_{1,i} (\boldsymbol{q}_i + F_i^T \boldsymbol{\lambda}_1) \\
& + \sum_i s^4_i \boldsymbol{\lambda}_{2,i} (\boldsymbol{q}_i + F_i^T \boldsymbol{\lambda}_2),
\end{align*}
where $p^k_{i}, p^k_{i,j}$ are non-negative variables (sum-of-squares polynomials), $ s^k_i$ are free variables (polynomials with no restriction).
}

\section*{Acknowledgment}

We are grateful to the reviewers who have offered many constructive comments that have significantly improved the manuscript. We also thank Brian Acosta, William Yang and Yu-Ming Chen for their help with the experimental setup.

\bibliographystyle{ieeetr}
\bibliography{Refs}

\vspace{-20 mm}
\begin{IEEEbiography}[{\includegraphics[width=1in,height=1.25in,clip,keepaspectratio]{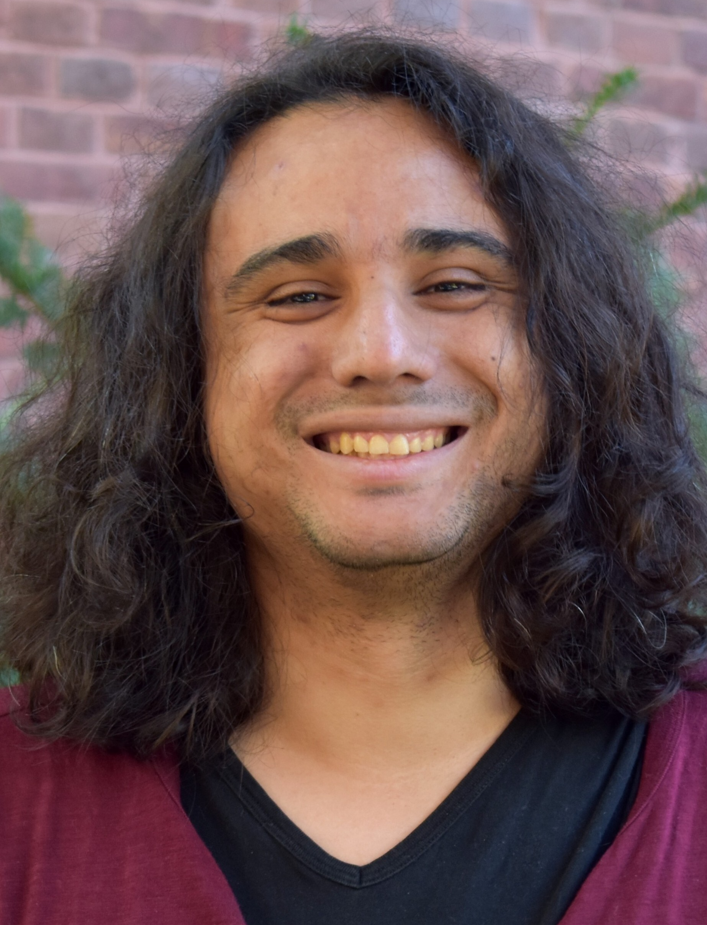}}]%
	{Alp Aydinoglu}
	completed his B.S. in Control Engineering from Istanbul Technical University in 2017 and is currently pursuing his Ph.D. in Electrical and Systems Engineering at University of Pennsylvania, working with Michael Posa in Dynamic Autonomy and Intelligent Robotics (DAIR) Lab. His research emphasizes control of multi-contact systems.
\end{IEEEbiography}
\vspace{-14 mm}
\begin{IEEEbiography}[{\includegraphics[width=1in,height=1.25in,clip,keepaspectratio]{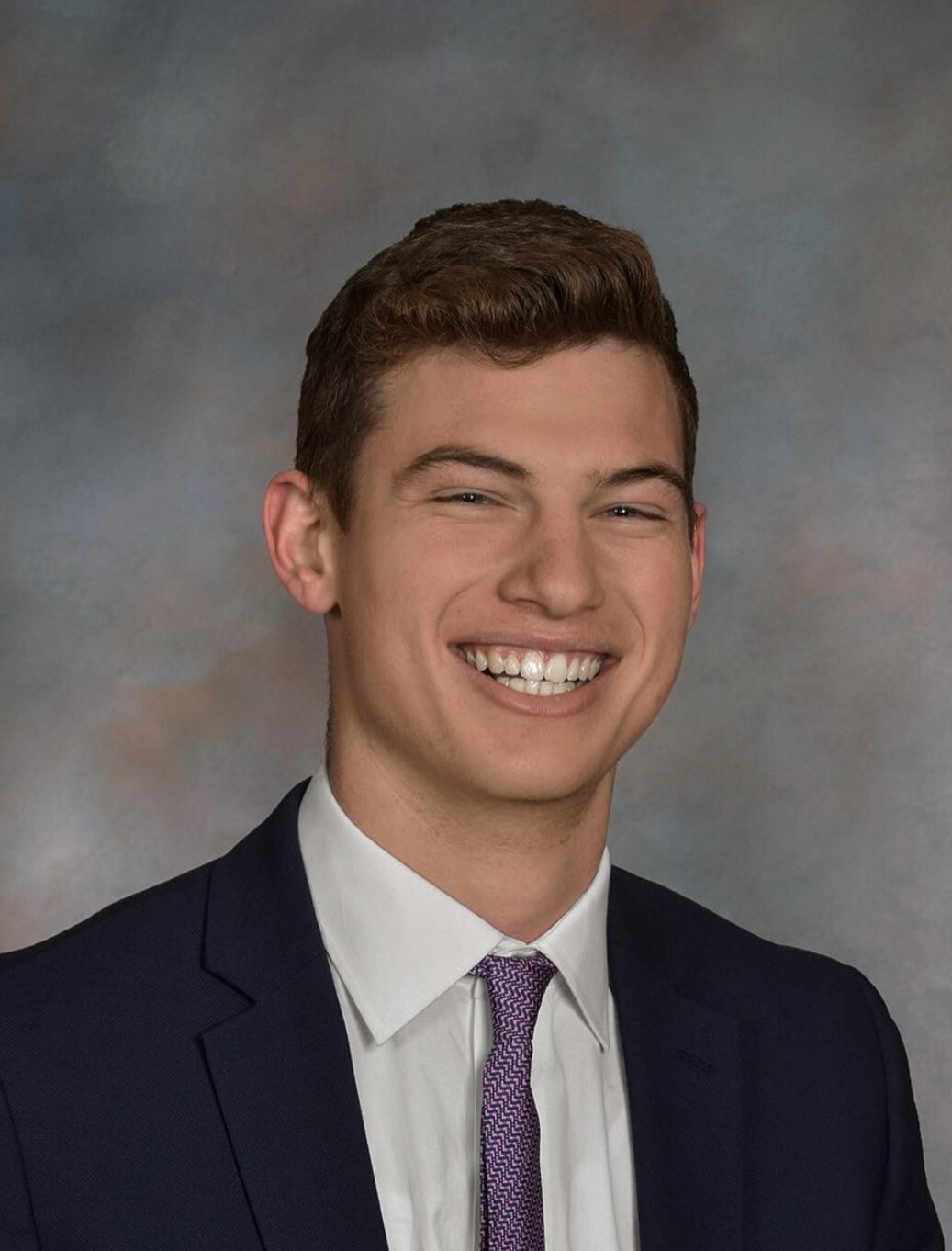}}]%
	{Philip Sieg}
	is currently working towards a B. Eng. degree in Mechanical Engineering and an MSE in Robotics at the University of Pennsylvania.
	His research interests include hardware design for robotics, legged locomotion, mobile robots, and grasping.
\end{IEEEbiography}
\vspace{-105 mm}
\begin{IEEEbiography}[{\includegraphics[width=1in,height=1.25in,clip,keepaspectratio]{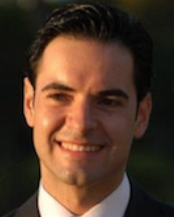}}]%
	{Victor M. Preciado}
	is an Associate Professor and Graduate Chair in the Department of Electrical and Systems Engineering at the University of Pennsylvania, where he is affiliated with the Networked and Social Systems Engineering program, the Warren Center for Network and Data Sciences, and the Applied Math and Computational Science program. He received the Ph.D. degree in Electrical Engineering and Computer Science from the Massachusetts Institute of Technology and was a postdoctoral researcher at the GRASP lab. He was a recipient of the 2017 National Science Foundation CAREER Award, the 2018 Best Paper Award by the IEEE Control Systems Magazine, and a runner-up of the 2019 Best Paper Award by the IEEE Transactions on Network Science and Engineering. He is an IEEE Senior Member, as well as Associate Editor of the IEEE Transactions on Network Science and Engineering and the IEEE Transactions on Control of Networked Systems.
\end{IEEEbiography}
\vspace{-107 mm}
\begin{IEEEbiography}[{\includegraphics[width=1in,height=1.25in,clip,keepaspectratio]{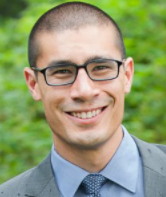}}]%
	{Michael Posa}
	received the Ph.D. degree in Electrical Engineering and Computer Science from the Massachusetts Institute of Technology, in 2017. He is currently an Assistant Professor of Mechanical Engineering and Applied Mechanics at the University of Pennsylvania, where he is a member of the General Robotics, Automation, Sensing and Perception (GRASP) Lab. He holds secondary appointments in Electrical and Systems Engineering and in Computer and Information Science. He leads to Dynamic Autonomy and Intelligent Robotics (DAIR) Lab, which focuses on developing computationally tractable algorithms to enable robots to operate both dynamically and safely as they maneuver through and interact with their environments, with applications including legged locomotion and manipulation. He has received the best paper award at Hybrid Systems: Computation and Control, a Google Faculty Research Award, and the Young Faculty Researcher award from the Toyota Research Institute. He is a member of IEEE and Associate Editor at IEEE Robotics and Automation Letters.
\end{IEEEbiography}

\end{document}